\documentclass{article} 
\usepackage{iclr2027_conference,times}

\usepackage{amsmath,amsfonts,bm}

\def\eqref#1{equation~\ref{#1}}

\def\1{\bm{1}}

\DeclareMathAlphabet{\mathsfit}{\encodingdefault}{\sfdefault}{m}{sl}
\SetMathAlphabet{\mathsfit}{bold}{\encodingdefault}{\sfdefault}{bx}{n}

\usepackage{hyperref}
\usepackage{url}
\usepackage{booktabs}
\usepackage{multirow}
\usepackage{amsmath}
\usepackage{amssymb}
\usepackage{bbding}
\usepackage{arydshln}
\usepackage{wrapfig}
\usepackage{enumitem}
\usepackage{tabularx}
\usepackage{makecell}
\usepackage{array}
\usepackage{subcaption}
\newcolumntype{Y}{>{\raggedright\arraybackslash}X}

\newcolumntype{C}[1]{>{\centering\arraybackslash}m{#1}}
\newcolumntype{L}[1]{>{\raggedright\arraybackslash}m{#1}}
\usepackage{xcolor}         
\definecolor{ao}{rgb}{0.0, 0.5, 0.0}
\definecolor{forestgreen}{RGB}{0, 150, 0}
\definecolor{lightgraybox}{RGB}{240,240,240}

\usepackage{inconsolata} 
\usepackage{ragged2e}

\usepackage[most]{tcolorbox}
\newtcolorbox{promptbox}[1]{
  enhanced,
  breakable,
  colback=white,
  colframe=black,
  boxrule=1pt,
  arc=3pt,
  left=10pt,
  right=10pt,
  top=8pt,
  bottom=8pt,
  title={#1},
  coltitle=white,
  colbacktitle=black,
  fonttitle=\bfseries\rmfamily,
  boxed title style={
    colback=black,
    colframe=black,
    sharp corners
  }
}

\title{Interpretability Is Not Just a Tax: Sparsity and Diversity Yield Legible and Composable Reward Experts for Personalization}

\author{\textbf{Yifan Wang\textsuperscript{1},}
        \textbf{Jinyi Mu\textsuperscript{2},}
        \textbf{Mayank Jobanputra\textsuperscript{1},}
        \textbf{Yu Wang\textsuperscript{3},}
        \\
        \textbf{Ji-Ung Lee\textsuperscript{1}, }
        \textbf{Soyoung Oh\textsuperscript{1}, }
        \textbf{Isabel Valera\textsuperscript{1,4}, }
        \textbf{Vera Demberg\textsuperscript{1,5}} 
        \vspace{0.6em}
        \\
        \textsuperscript{1} Saarland University  \space \textsuperscript{2} University of California San Diego \space
        \textsuperscript{3} Bielefeld University \\
        \textsuperscript{4} Max Planck Institute for Software Systems \space
        \textsuperscript{5} Max Planck Institute for Informatics \vspace{0.4em} \\
        \texttt{yifwang@lst.uni-saarland.de}
}

\iclrfinalcopy 
\begin{document}

\maketitle
\fancyhead{}

\begin{abstract}

Preference modeling aims to capture heterogeneous human values and increasingly relies on learning latent components to represent this variation without fine-grained supervision.
What these components actually represent, however, is rarely examined.
We audit Mixture-of-Experts (MoE) reward models along three rungs---semantic coherence, functional correspondence, and composability---in controlled and real-world settings, holding the architecture and training data fixed so that differences isolate the effect of the training objective.
Our diagnosis is that the components conventional MoE reward models learn are entangled, which limits not only interpretability but also the personalization such models are built for.
The entanglement is not intrinsic to the architecture but a property of the training objective: off-the-shelf sparsity and diversity regularizers already substantially reduce it.
The resulting configuration, which we call sparse MoE, recovers all underlying semantic categories without domain supervision in a controlled setting, reaching 84.8\% expert purity against 60.8\% for conventional MoE; on real-world data its experts admit human-interpretable task- and domain-level descriptions aligned with their functional strengths.
Targeted interventions on routing weights produce predictable preference changes, showing that these interpretations are behaviorally grounded rather than merely descriptive.
Crucially, the disentanglement induced by this objective also makes the experts composable: lightweight router updates yield a 25.81-point improvement in few-shot personalization with 50 examples, roughly 2.7 times the gain of the identical architecture and adaptation procedure applied to entangled experts.
Together, these results show that MoE reward models can learn interpretable expert structure, and that the same structure can substantially improve personalization.

\end{abstract}

\section{Introduction}



Reinforcement learning from human feedback (RLHF) commonly relies on reward models to capture human preferences and align large language models (LLMs) with desired behavior~\citep{finetune-rlhf, rlhf-bai, gpt4, rlhf-survey}.
Most reward models, however, represent preferences with a single global function, despite human preferences being heterogeneous and context dependent~\citep{reward-model-survey-2025-emnlp}.
While earlier approaches model such heterogeneity using fine-grained annotations of preference dimensions, recent work learns multiple latent preference components directly from binary preference data without such supervision and combines them to represent diverse or individualized preferences~\citep{test-time-alignment-2024, luo-etal-2025-rethinking, shen-etal-2025-micro}.
This line of work, however, primarily emphasizes predictive or personalization performance, leaving open whether the learned components correspond to coherent and human-interpretable structure.

\begin{figure*}[ht!]
    \centering
    \includegraphics[width=\linewidth]{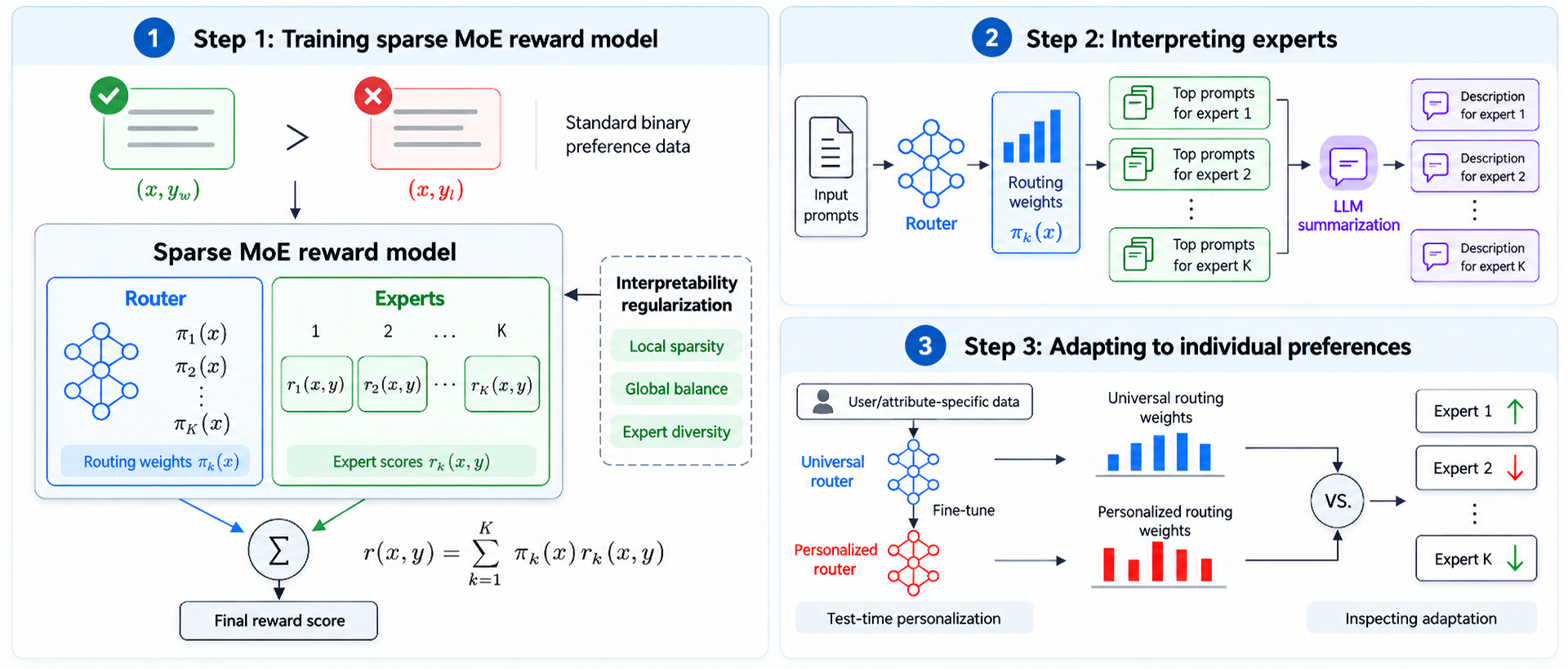}
    \caption{\textbf{The models we study and how we analyze them: training an MoE reward model, interpreting its experts, adapting it to individual preferences, and inspecting the adaptation.}
    An MoE reward model is trained on standard binary preference data with interpretability regularization, and its experts can be interpreted as the domains in which they are activated and specialize.
    At test time, the router can be fine-tuned to fit personalized preferences, and changes in routing weights show how the model adapts.
    }
    \label{fig:workflow}
    \vspace{-1.0em}
\end{figure*}

To address this gap, we ask whether latent components in preference models can become \textbf{human-interpretable and functionally specialized}, rather than merely providing additional representational capacity.
MoE reward models are a natural setting for this question: their routers expose \emph{where} individual experts are activated, while the experts themselves can be evaluated to determine \emph{what} they are good at~\citep{olson-etal-2025-probing}.
Across controlled and real-world experiments, we find that entanglement is the default and that it is costly: conventional MoE reward models route almost uniformly and learn experts that behave almost identically, whereas sparsity and diversity regularizers induce experts whose activation patterns reveal coherent, prompt-identifiable domains and tasks and whose functional strengths align with these interpretations. 
To establish this, we evaluate latent preference components along three rungs of increasing strength: whether routing patterns are semantically \emph{coherent}, whether they \emph{correspond} to expert strength and behavior, and whether the disentangled components can be \emph{composed} into new preference functions through router updates.
We hold the architecture of~\citet{shen-etal-2025-micro}, the backbone, and the training data fixed and vary only the objective, isolating it as the cause of the structure we observe. 
Figure~\ref{fig:workflow} summarizes the pipeline: discovering, validating, and composing expert structure.

Our contributions are threefold.
\textbf{First}, we introduce a three-rung protocol for auditing latent preference components through coherence, correspondence, and composability.
\textbf{Second}, we show that conventional MoE reward models learn highly entangled experts, while sparsity and diversity regularization produces coherent, functionally specialized experts: in a controlled setting, expert purity rises from 60.8\% to 84.8\% and specialization from $-0.43$ to $5.56$, with similar gains also on real-world preference data. Targeted interventions further confirm that these interpretations reflect genuine expert behavior.
\textbf{Third}, we show that this specialization also improves personalization: with 50 adaptation examples, sparse MoE gains 25.81 points on RPR, compared with 9.57 points for the same architecture without regularization. Thus, the same specialization that makes experts interpretable also makes them more composable for personalization.

\section{Related Work}
\label{sec:related work}

\paragraph{Latent Preference Modeling and Personalization}
Most existing reward models are trained to reflect universal human values, yet they often fail to capture their heterogeneous nature and are therefore not well-suited for representing individualized preferences~\citep{maxmin-rlhf-2024, zhang2025diverging}.
Many recent approaches introduce multiple preference components with additional supervision, such as fine-grained attribute annotations~\citep{wang-etal-2024-arithmetic, quan-2024-dmoerm, zhou-etal-2024-beyond, wu2026reward} or user-specific histories~\citep{zhao2024group, choi-etal-2025-copl, liu-etal-2025-llms, chen2025pal}, and represent unseen preferences at test time by reweighting these predefined attributes or prototypes. While effective, they depend on costly extra annotation.

To reduce this requirement, recent work learns latent preference structure directly from binary preference data.
Some approaches introduce latent user variables, via variational autoencoders~\citep{poddar-vae-preference-2024} or expectation-maximization over user prototypes~\citep{chidambaram2026direct}.
Others learn multiple reward components and adapt their mixture, through ensembles of reward models~\citep{test-time-alignment-2024} or principal component analysis of response representations~\citep{luo-etal-2025-rethinking}.
Closest to our work is MiCRo~\citep{shen-etal-2025-micro}, which trains an MoE reward model on standard preference data and adapts its router to context-dependent preferences.
These approaches show that latent components improve preference modeling, but leave open what the components represent: whether they are coherent, human-interpretable, and aligned with distinct functional roles.

\paragraph{Interpretability through Sparsity and MoE}

Prior work has shown that sparsity can help disentangle LLM representations and recover more interpretable units~\citep{elhage2022superposition, bricken2023monosemanticity}.
Sparse autoencoders (SAEs), for example, project representations into a higher-dimensional space under sparsity constraints, yielding monosemantic features that are interpreted by summarizing highly activating examples~\citep{shu-etal-2025-sae-survey}, and these features have been used to build interpretable reward functions~\citep{movva2026whats, sparserm, saerm}.
Such pipelines learn features separately from the reward objective, so the features are not necessarily preference-relevant, and their weights are typically fixed across inputs, limiting flexible preference modeling.

In parallel, MoE architectures have emerged as an effective paradigm for scaling LLMs~\citep{moe-survey-2025}.
Their explicit routing structure also provides a natural lens for interpretability, as routing patterns often align with semantic structures, and experts exhibit emergent specialization~\citep{fayyaz2026steering}.
MoE reward models are therefore a natural setting for studying latent preference components: unlike SAE-based pipelines, their expert structure is learned jointly with the reward objective, and their context-dependent routing allows experts to be recombined across inputs.
Yet beyond~\citet{wang-etal-2024-interpretable} and~\citet{shen-etal-2025-micro} they remain largely underexplored, and scalar preference supervision provides no explicit semantic signal for how latent experts should organize.

Following~\citet{do-etal-2026-domain}, we interpret experts as the domains and contexts in which they are preferentially activated and exhibit specialization, such as math problem solving or feedback seeking, and ask whether such structure can emerge from binary preference supervision alone.
This notion of interpretation identifies \emph{where} an expert is activated and functionally specialized. 
Such structure is naturally useful for personalization: if experts specialize in different contexts, reweighting and recombining them is a natural way to emphasize different behaviors for different users. It also enables targeted control and auditing of how reward models organize and adapt their preferences.

\section{MoE Reward Models and Interpretability Regularization}
\label{sec:losses}
\paragraph{Preliminary} 
In RLHF, a reward model $r_\theta(x, y) \in \mathbb{R}$ estimates human preferences over responses $y$ given a prompt $x$, and is trained on pairwise comparisons $\mathcal{D} = \{(x, y_w, y_l)\}$ in which $y_w$ is preferred over $y_l$.
It is typically a Bradley--Terry model~\citep{bradley1952rank}, which defines the preference probability as:
\begin{equation}
P(y_w \succ y_l \mid x)  = \sigma\big(r_\theta(x, y_w) - r_\theta(x, y_l)\big),
\end{equation}
where $\sigma(z) = (1 + e^{-z})^{-1}$.
The model is trained by minimizing the negative log-likelihood of observed preferences.

\paragraph{MoE Reward Model} 
\citet{wang-etal-2024-interpretable} and \citet{shen-etal-2025-micro} propose using MoE architectures to model diverse and heterogeneous preferences.
Specifically, an MoE reward model consists of $K$ experts $\{r_{\theta_k}\}_{k=1}^K$ and a router $\pi_\phi(x) \in \Delta^{K-1}$.
In practice, we implement experts as linear heads and the router as a one-hidden-layer MLP over backbone model representations.
The router produces a softmax distribution over experts based on the input $x$, where $\pi_{\phi,k}(x)$ denotes the weight assigned to expert $k$.
The final reward is computed as the weighted average of expert scores:
\begin{equation}
r(x, y) = \sum_{k=1}^K  \pi_{\phi,k}(x)\, \cdot r_{\theta_k}(x, y).
\end{equation}
During training, we optimize the marginalized expert-level BT loss following \citet{shen-etal-2025-micro}: 
\begin{equation}
\begin{aligned}
\mathcal L_{\mathrm{RM}}
=
-\frac{1}{|\mathcal D|}
\sum_{(x,y_w,y_l)\in\mathcal D}
\log \sum_{k=1}^K
\big(\pi_{\phi,k}(x) \cdot \sigma\!(r_{\theta_k}(x,y_w) 
 - r_{\theta_k}(x,y_l))\big).
\end{aligned}
\end{equation}

Compared to applying BT loss to the expected expert score, the marginalized expert-level loss assigns comparison-level gradients to individual experts rather than broadcasting the same preference signal to all active experts, encouraging more diverse expert behaviors.
\paragraph{Interpretability Regularization} To test whether the entanglement of conventional MoE stems from the architecture or from the training objective, we hold the architecture fixed and add three standard interpretability regularizers to the objective, encouraging disentangled routing and the emergence of expert specialization; we refer to this configuration as sparse MoE.
Using established regularizers keeps the manipulation minimal and interpretable: structure that emerges can be attributed to the objective itself rather than to the particular mechanism that implements it.

\noindent\textbf{(1) Local Sparsity}: 
Following prior work on interpretability through sparsity~\citep{elhage2022superposition, bricken2023monosemanticity}, we encourage sparse routing distributions by minimizing the entropy of the routing weights:
\begin{equation}
\begin{aligned}
\mathcal L_{\mathrm{ls}}
=
\mathbb{E}_{x \sim \mathcal D}
\left[
\frac{H\!\big(\pi_\phi(x)\big)}{\log K}
\right], H(\pi) = -\sum_{k=1}^K \pi_k \log \pi_k.
\end{aligned}
\label{eq:local entropy}
\end{equation}
\noindent\textbf{(2) Global Balance}: To prevent model collapse, we encourage balanced utilization across experts by maximizing the entropy of the average routing distribution within a batch $\mathcal B$:
\begin{equation}
\begin{aligned}
\mathcal L_{\mathrm{gb}}
= 
\mathbb{E}_{\mathcal B \sim \mathcal D}
\left[
- \frac{H(\bar{\pi}_\phi)}{\log K}
\right],
\bar{\pi}_\phi = \mathbb{E}_{x \sim \mathcal B}[\pi_\phi(x)].
\end{aligned}
\end{equation}
\noindent\textbf{(3) Expert Diversity}: We encourage experts to capture distinct preference signals by penalizing correlations between their reward differences. 
Let $\Delta_k = r_{\theta_k}(x,y_w) - r_{\theta_k}(x,y_l)$,
and we minimize the average squared pairwise Pearson correlation within a batch $\mathcal B$ of preference pairs:
\begin{equation}
\mathcal L_{\mathrm{div}}
=
\frac{2}{K(K-1)}
\sum_{i<j}
\mathrm{corr}(\Delta_i, \Delta_j)^2.
\label{eq:expert correlation}
\end{equation}
The overall training objective is defined as $\mathcal L
=
\mathcal L_{\mathrm{RM}}
+ \lambda_{\mathrm{ls}} \mathcal L_{\mathrm{ls}}
+ \lambda_{\mathrm{gb}} \mathcal L_{\mathrm{gb}}
+ \lambda_{\mathrm{div}} \mathcal L_{\mathrm{div}}$, 
where $\lambda_{\mathrm{ls}}$, $\lambda_{\mathrm{gb}}$, and $\lambda_{\mathrm{div}}$ control the strengths of local sparsity, global balance, and expert diversity.

\section{Evaluation Protocol: Coherence, Correspondence, and Composability}
\label{sec:protocol}

We evaluate whether latent preference components are interpretable along three rungs, each answering a question the previous one leaves open.
\textit{Coherence} asks whether routing patterns admit a consistent semantic description, measured by category recovery and expert purity in the controlled setting (Section~\ref{sec:domain_retrieval}) and by description fidelity and human ratings on real-world data (Section~\ref{sec:real_interpretability}).
\textit{Correspondence} asks whether those descriptions track expert behavior, tested at two levels: \emph{correlationally}, by an \textit{expert specialization} score measuring whether an expert outperforms its peers on the inputs routed to it (Sections~\ref{sec:domain_retrieval} and~\ref{sec:real_interpretability}), and \emph{causally}, by targeted attribute steering (Section~\ref{sec:attribute_steering}).
\textit{Composability} asks whether disentangled components can be combined into new preference functions, measured by test-time preference adaptation (Section~\ref{sec:adaptation}).
Each rung rules out a weaker reading of the one below: coherence alone may reflect only a partition of the inputs; a correlational association may reflect a router that tracks input statistics rather than experts with distinct roles; and control over individual experts may still not extend to combining them for preferences the model was never trained on.

\paragraph{Models} Following~\citet{shen-etal-2025-micro}, we use a strong reward model \href{https://huggingface.co/Ray2333/GRM-Llama3.2-3B-rewardmodel-ft}{GRM-Llama3.2-3B}~\citep{yang2024grm} as the frozen backbone and train $K$ linear heads as experts. 
Unless otherwise specified, we set $\lambda_{\mathrm{ls}}$, $\lambda_{\mathrm{gb}}$, and $\lambda_{\mathrm{div}}$ to $0.5$, $0.5$, and $2.0$.
The number of experts is set per experiment according to the data complexity.
Appendix~\ref{app:experimental details} details the setups; Appendices~\ref{app:num_expert} and~\ref{app:ablation} ablate the expert number $K$ and the regularization terms and give guidelines for selecting hyperparameters; Appendix~\ref{app:generalization} repeats the real-world experiments with \href{https://huggingface.co/Skywork/Skywork-Reward-V2-Qwen3-8B}{Skywork-Reward-V2-Qwen3-8B} as the backbone.

\paragraph{Baselines} A single-head reward model serves as a reference for universal preference modeling, and among MoE models we compare vanilla MoE, MiCRo~\citep{shen-etal-2025-micro}, which encourages a uniform routing distribution, and sparse MoE.
In the real-world setting we add two SAE-based baselines, SARM~\citep{sparserm} and WIMHF~\citep{movva2026whats}; because their interpretation mechanisms differ, we exclude them from the human evaluation and report personalization only for SARM.
Appendix~\ref{app:moe vs. sae} compares MoE- and SAE-based reward models directly.

\section{Diagnosing Expert Structure: Controlled Settings}


We first consider two controlled settings in which the underlying structure is known, which lets us test the first two rungs of the protocol (Section~\ref{sec:protocol}) directly: whether the router recovers semantically coherent structure, whether the associated experts are strongest where they are routed, and whether intervening on them changes preferences as predicted.

\subsection{Recovering Latent Semantic Structure}
\label{sec:domain_retrieval}

We investigate whether MoE can discover latent semantic structure in the data without domain supervision.
We use the \href{https://huggingface.co/datasets/stanfordnlp/SHP}{Stanford Human Preferences} (SHP;~\citealp{shp}) dataset, which includes annotations of the prompt domains.
As these domains vary in granularity, we group the 16 subsets into four categories: \textit{academic}, \textit{professional}, \textit{cooking}, and \textit{fiction}.

We train MoE reward models with $K=5$ experts on the dataset without using the domain annotations.
After training, we assign each expert a category label according to the majority category among its 50 highest-routed validation examples.
We then evaluate (i) whether routing patterns recover the held-out category structure and (ii) whether experts are functionally specialized in the categories to which they are preferentially routed.

\paragraph{Evaluation Setting} In addition to the task accuracy, we report two model behavior metrics: average routing entropy (Eq.~\ref{eq:local entropy}) and average expert correlation (reported without squaring; Eq.~\ref{eq:expert correlation} uses its squared form for training).
Values closer to zero indicate sparser routing and more distinct expert behavior, respectively, which are desirable for routing-pattern interpretability and expert specialization.
Since the latent categories are known, we further evaluate interpretability and specialization directly using the following metrics adapted from prior work~\citep{dscp-moe-2025, yang2025mixture, expert_strikes}:
\begin{itemize}[leftmargin=1em]
    \item \textit{Category Coverage}: the proportion of categories assigned to at least one expert, measuring how much of the underlying structure the router captures.
    \item \textit{Expert Purity}: for each expert, the proportion of its 50 top-activating test examples that belong to its assigned category; higher values mean more coherent routing.
    \item \textit{Expert Specialization}: for each expert, the gap between its accuracy on its labeled category and the average accuracy of all experts on that category; higher values mean the expert is functionally stronger where it is routed.
\end{itemize}

\begin{table*}[ht!]
    \centering
    \caption{\textbf{Latent structure recovery and expert specialization on SHP.} 
    Without access to domain labels during training, sparse MoE recovers coherent routing structure and develops experts that are functionally specialized in their routed categories.
    Results are averaged over three runs.}
    
    \resizebox{\textwidth}{!}
    {
    \begin{tabular}{lcccccc}
    \toprule
    & \multicolumn{3}{c}{Model Behavior} & \multicolumn{3}{c}{Interpretable Structure} \\
     Method & Accuracy ($\uparrow$) & Routing Entropy  ($\downarrow$) & Expert Correlation  ($\to 0$) & Category Coverage ($\uparrow$) & Expert Purity ($\uparrow$) & Expert Specialization ($\uparrow$) \\
     \midrule
Single Reward & 70.71 & - & - & - & - & - \\
Vanilla MoE & \textbf{72.88} & 0.996 & 0.594 & 0.67 & 60.80 & -0.43 \\
MiCRo & 72.64 & 1.000 & 0.525 & 0.42 & 50.27 & 0.24 \\

\midrule
Sparse MoE & 70.18 & \textbf{0.157} & \textbf{0.010} & \textbf{1.00} & \textbf{84.76} & \textbf{5.56} \\
    \bottomrule
    \end{tabular}
    }
    \vspace{-1.0em}
    \label{tab:domain retrieval}
\end{table*}

\paragraph{Results}
Table~\ref{tab:domain retrieval} shows a clear difference in the structure learned by the three MoE variants.
Sparse MoE is the only model that recovers all four latent categories, despite receiving no category supervision during training.
By contrast, vanilla MoE and MiCRo exhibit nearly uniform routing and reach category coverage of only 0.67 and 0.42.
Sparse MoE's routing is also far more coherent: 84.76\% of each expert's highest-routed examples belong to that expert's assigned category, against 60.80\% for vanilla MoE and 50.27\% for MiCRo.
Crucially, this structure comes with distinct expert behavior: sparse MoE's experts beat the average expert by 5.56 points on their assigned categories, whereas vanilla MoE's and MiCRo's match or trail it.
The discovered categories are therefore not merely patterns in input assignment; they correspond to experts with different functional strengths.
Regarding task performance, sparse MoE achieves 70.18\% overall accuracy, compared with 72.88\% for vanilla MoE and 72.64\% for MiCRo.
Thus, the stronger structural organization comes with a modest 2--3 point reduction in aggregate accuracy.
The drop is concentrated in the \textit{academic} and \textit{professional} categories, which are broad and heterogeneous: there the strongest expert for an input often does not receive the highest routing weight (Appendix~\ref{app:complete_results}).
The cost therefore reflects imperfect routing rather than a limit on what the experts themselves can represent.
\begin{table*}[ht!]
    \centering

    \caption{\textbf{Input prompts together with the sparse MoE routing weights assigned to each expert.}
    Parentheses under each expert indicate the category associated with that expert.
    The router produces sparse and semantically meaningful routing distributions based on the input semantics.}
    \resizebox{\linewidth}{!}{
    \begin{tabular}{L{0.9\textwidth} C{1.6cm} C{1.6cm} C{1.6cm} C{1.6cm} C{1.6cm} C{1.2cm}}
    \toprule
    Input Prompt & Category &\makecell{Expert 0\\(Academic)} &\makecell{Expert 1\\(Academic)} &\makecell{Expert 2\\(Professional)} &\makecell{Expert 3\\(Cooking)} &\makecell{Expert 4\\(Fiction)} \\
    \midrule

    Why is 18 the maximum amount of electrons an atomic shell can hold?
    & Academic & 0.025 & \textbf{0.967} & 0.000 & 0.007 & 0.000 \\

    \midrule

    Why is it that Eastern and Western philosophy don't appear to have interacted much, even in the ancient world, until the 19th century? Is there any evidence of cross pollination of ideas?
    & Academic & \textbf{0.777} & 0.222 & 0.000 & 0.000 & 0.001 \\

    \midrule

    I own a condo and my downstairs neighbors have called police on me over 300 times in the last 365 days claiming frivolous noise complaints, HOA is now involved and threatening legal action. Can they force me out of my unit? No mortgage, bought in cash...I am sole occupant. (MN)
    & Professional & 0.000 & 0.000 & \textbf{0.989} & 0.000 & 0.011 \\

    \midrule

    A friend said I should crack my garlic 10 minutes before cooking, is there a real reason for this? She said that it causes some chemical reaction to make the garlic taste better, can I just cut the garlic up and wait 10 minutes or is this pointless to begin with?
    & Cooking & 0.000 & 0.123 & 0.000 & \textbf{0.877} & 0.000 \\

    \midrule

    [Marvel Comics] Why did Norse paganism fade away, when Thor was actually present on Earth, fighting monsters and performing miracles?
    & Fiction & 0.178 & 0.000 & 0.000 & 0.000 & \textbf{0.822} \\

    \bottomrule
    \end{tabular}
    }    
    \label{tab:qualitative_examples}
    \vspace{-1.0em}
\end{table*}

\paragraph{Qualitative Analysis}

Table~\ref{tab:qualitative_examples} shows representative SHP prompts together with their sparse MoE routing distributions.
The router typically assigns most probability mass to a single expert, and the dominant expert is consistent with the semantic category inferred from that expert's highest-routed examples.
Routing need not be strictly one-hot: some prompts assign non-negligible mass to a secondary expert when an input involves multiple semantic aspects, suggesting that the model combines several learned components when appropriate.

The learned structure can also be finer-grained than the four evaluation categories.
Within the broad \textit{academic} category, the science question is routed primarily to Expert~1, whereas the humanities question is routed primarily to Expert~0.
Although both receive the same coarse category label, their routing behavior suggests that the model has discovered finer semantic distinctions than the evaluation labels represent.


\vspace{-0.5em}
\begin{center}
\colorbox{gray!15}{%
  \parbox{0.98\linewidth}{\textbf{Takeaway (coherence and correspondence)}:
  Without domain supervision, sparse MoE recovers human-recognizable latent structure from binary preference data, and the resulting routing patterns align with distinct expert strengths.}}
\end{center}
\vspace{-0.5em}

\subsection{Functional Validation through Targeted Attribute Steering}
\label{sec:attribute_steering}

The correspondence established so far is correlational: experts are strongest on the inputs routed to them.
We now test it causally, through targeted post-hoc interventions on routing weights in a controlled setting with known preference attributes.
If routing patterns identify genuine expert roles, intervening on the corresponding experts should produce predictable and targeted changes in model behavior.


We train MoE models with $K=10$ experts on the \href{https://huggingface.co/datasets/microsoft/rpr}{Reasonable Preference Reversal} (RPR;~\citealp{rpr}) dataset, which spans 40 attributes and whose relative preference reverses under each (e.g., clarity vs.\ creativity).
Following~\citet{shen-etal-2025-micro}, we append a specification of the desired attribute to each prompt during training so the model learns attribute-conditioned preferences.
The attribute enters only as input context: the objective remains the binary preference loss, no expert is assigned an attribute, and the expert--attribute associations used below are derived post hoc from routing statistics.
At test time we remove that specification, so preference changes are induced solely by interventions on the router.
For each attribute we select the target expert whose top 100 activating validation examples contain the highest proportion of that attribute---our interpretation of the expert's role---set its weight to $w \in \{0.25, 0.5, 0.75, 1.0\}$, and rescale the remaining weights to keep the distribution normalized.
A further baseline selects target experts at random.

\paragraph{Evaluation Setting}

We evaluate steering effectiveness using two metrics: \textit{desired flip rate} and \textit{undesired flip rate}.
For each attribute, we first identify the sets of examples that are correctly and incorrectly classified when the target expert is removed.
The desired flip rate is defined as the proportion of initially incorrect examples that switch to the correct response after assigning a specified weight to the target expert.
Conversely, the undesired flip rate measures the proportion of initially correct examples that become incorrect after manipulating the target expert's weight.
Intuitively, effective and targeted steering should achieve a high desired flip rate without increasing the undesired flip rate.
The final scores are averaged over all target weights and attributes.

\begin{figure}[t]
    \centering

    \begin{subfigure}[t]{0.48\linewidth}
        \centering
        \includegraphics[width=\linewidth]{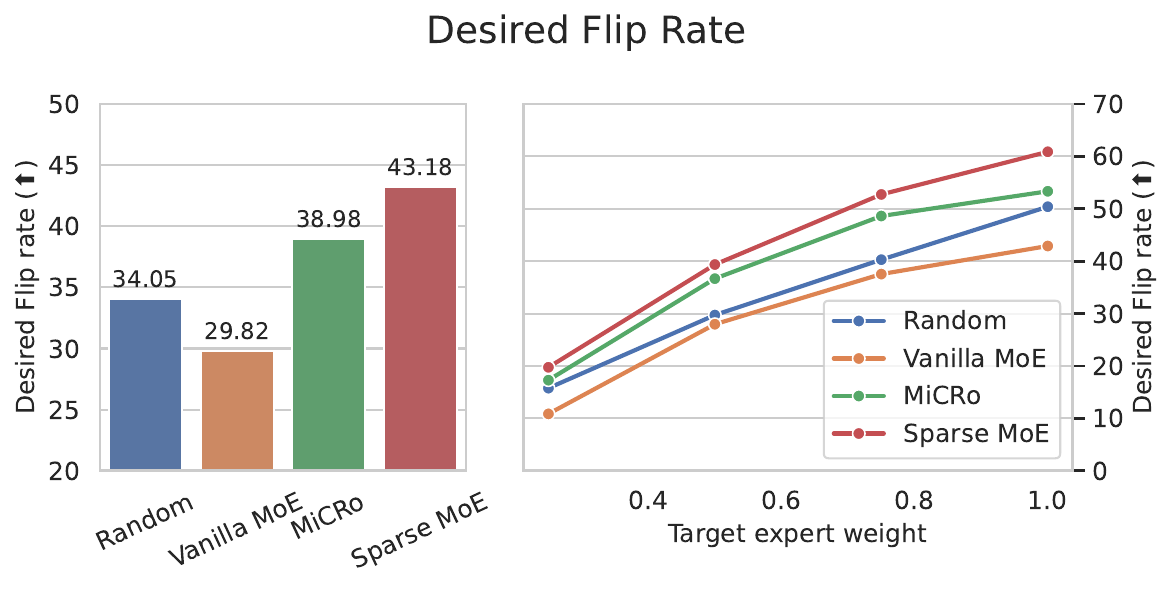}
    \end{subfigure}
    \hfill
    \begin{subfigure}[t]{0.48\linewidth}
        \centering
        \includegraphics[width=\linewidth]{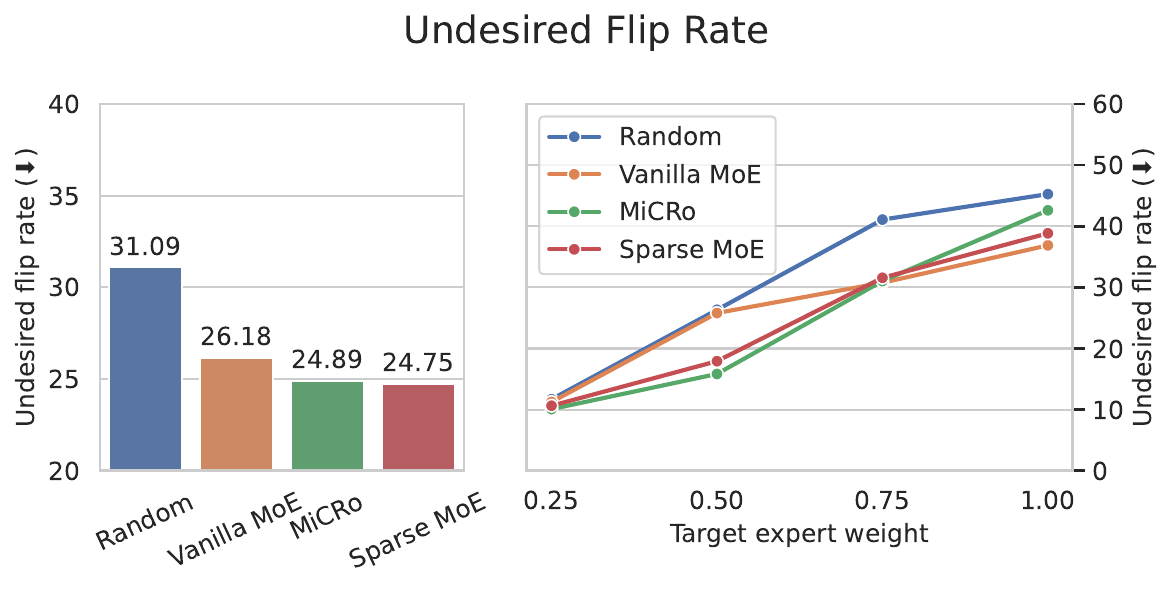}
    \end{subfigure}

    \caption{\textbf{Attribute steering results on RPR.}
    Sparse MoE achieves the highest desired flip rate (left) while keeping a low undesired flip rate (right), showing strong and targeted preference steering.
    Results are averaged over three runs.}
    \vspace{-1.0em}
    \label{fig:flip_rate}
\end{figure}

\paragraph{Results}

Figure~\ref{fig:flip_rate} shows the desired and undesired flip rates both on average and as a function of the target expert weight.
Sparse MoE achieves the highest desired flip rate, outperforming the strongest baseline by 4.20 points on average and exceeding 60\% when the target expert receives full routing weight ($w=1.0$).
The rate also rises consistently as the target expert is upweighted, so intervening on routing-identified experts produces predictable shifts toward the corresponding attributes.
Two effects contribute. First, sparse MoE's experts are behaviorally distinct, which makes routing interventions more influential in general: even randomly selected experts produce substantial preference changes when upweighted.
Second, selecting experts by their routing-based attribute associations improves the desired flip rate further, indicating that the semantic patterns revealed by routing carry real information about experts' roles.
Sparse MoE also keeps a low undesired flip rate, so its interventions are targeted rather than indiscriminate, whereas vanilla MoE steers substantially more weakly, consistent with its limited specialization, and MiCRo falls in between.
The routing patterns learned by sparse MoE are therefore behaviorally grounded: experts identified from their activations can be manipulated to change preferences predictably.

\vspace{-0.5em}
\begin{center}
\colorbox{gray!15}{%
  \parbox{0.98\linewidth}{\textbf{Takeaway (causal correspondence)}: The semantic structure recovered by sparse MoE is not merely descriptive; routing-based expert interpretations identify functionally distinct components whose targeted intervention predictably changes model preferences.}} 

\end{center}
\vspace{-0.5em}

\section{Diagnosing Expert Structure: Real-World Settings}

Having established coherence and correspondence where the latent structure is known, we ask whether the same structure emerges in real-world preference data, where no ground-truth decomposition exists.
We train and interpret MoE models on \href{https://huggingface.co/datasets/hendrydong/preference_700K}{preference-700K} (700K;~\citealp{dong2024rlhf}), using $K=20$ experts and $\lambda_{\mathrm{gb}}=1.0$ for sparse MoE to capture diverse patterns without routing collapse; Appendix~\ref{app:generalization} reports the same analyses for \href{https://huggingface.co/datasets/lmarena-ai/arena-human-preference-140k}{LMArena}.

\subsection{Human-Interpretable and Functionally Specialized Experts}
\label{sec:real_interpretability}
Following prior work~\citep{shu-etal-2025-sae-survey}, we interpret each expert by summarizing its top-activating examples: we collect the 20 validation prompts with the largest routing weights and prompt an LLM (\href{https://huggingface.co/Qwen/Qwen3.6-27B}{Qwen3.6-27B}) to describe their shared characteristics in natural language.
This description is our routing-based interpretation of the expert; Table~\ref{tab:expert_interpretation} shows examples for sparse MoE.

\begin{table*}[h]
    \centering
    \caption{\textbf{Natural language descriptions of experts in sparse MoE.}
    Highest-routed examples reveal topic- and domain-level patterns (all descriptions in Appendix~\ref{app:complete_results}).
    }
    \resizebox{\linewidth}{!}
    {
    \begin{tabular}{lccc}
    \toprule
     ID & Natural Language Description & Description Fidelity ($\uparrow$) & Expert Specialization ($\uparrow$)  \\ 
     \midrule
    0 & asks for advice, feedback, or discussion on a personal, professional, or academic challenge & 0.43 & 22.08 \\
    6 & asks to solve a procedural task involving numerical or list manipulation & 0.51 & 25.98 \\
    8 & asks for a summary or extraction of information from a provided text & 0.45 & 19.35 \\
    10 & asks for a direct answer to a multiple-choice or short-answer question & 0.36 & 20.56 \\
    13 & requests assistance with illegal, harmful, or sexually explicit activities & 0.62 & 3.35 \\
    15 & asks to solve a competitive mathematics problem & 0.42 & 32.00 \\
    
    \bottomrule
    \end{tabular}
    }
    
    \label{tab:expert_interpretation}
    \vspace{-1.0em}
\end{table*}

\paragraph{Evaluation Setting}
Without ground-truth expert labels, we test coherence and correlational correspondence with two metrics.
For routing pattern interpretability, we measure \textit{description fidelity} following~\citet{movva2026whats}.
For each expert, an LLM (\href{https://huggingface.co/Qwen/Qwen3.6-35B-A3B}{Qwen3.6-35B-A3B}) judges whether 200 test examples match its natural-language description, and we compute the Pearson correlation between these judgments and the corresponding routing weights.
Higher fidelity indicates that routing patterns are coherent and well captured by the expert description.
For functional specialization, we adapt the \textit{expert specialization} score from Section~\ref{sec:domain_retrieval}.
For each expert, we compare its accuracy on its top 10\% highest-routed examples with the average accuracy of all experts on the same examples.
A higher score indicates that the expert is functionally stronger where it is preferentially routed.

\paragraph{Results}

\begin{wraptable}{r}{0.48\linewidth}
    \centering
    \vspace{-4mm}
    \caption{\textbf{Semantic coherence and functional specialization.}
    Sparse MoE performs the best in both while preserving strong accuracy.}
    \resizebox{\linewidth}{!}
    {
    \begin{tabular}{lccc}
    \toprule
     Method & Accuracy ($\uparrow$) & Description Fidelity ($\uparrow$)  & Expert Specialization ($\uparrow$)  \\
     \midrule
Single Reward & 83.52 & - & -  \\
SARM & 75.72 & 0.09 & 2.48 \\
WIMHF & 60.43 & 0.21 & -9.06 \\
Vanilla MoE & \textbf{84.82} & 0.29 & 0.09 \\
MiCRo & 84.68 & 0.15 & 0.04  \\
Random & - & 0.01 & -0.04 \\
\midrule
Sparse MoE & 83.78 & \textbf{0.38} & \textbf{15.13}  \\
    \bottomrule
    \end{tabular}
    }
    
    \label{tab:interpretability}
    \vspace{-0.5em}
\end{wraptable}

Table~\ref{tab:interpretability} shows that sparse MoE learns substantially more coherent and functionally differentiated expert structure than the baselines.
Its description fidelity is the highest (0.38), indicating that its routing patterns are more consistently captured by natural-language interpretations.
More importantly, these semantic patterns align with distinct expert strengths: sparse MoE reaches an expert specialization score of 15.13, whereas vanilla MoE and MiCRo remain near zero.
The learned interpretations therefore describe not only where an expert is activated but also the contexts in which it performs particularly well, and Table~\ref{tab:expert_interpretation} shows that these contexts correspond to recognizable tasks and domains.
Sparse MoE retains competitive accuracy (83.78\%) against vanilla MoE (84.82\%) and MiCRo (84.68\%), while the SAE baselines are weaker on both axes.

\begin{wrapfigure}{r}{0.48\linewidth}
    \centering
    \vspace{-4mm}
    \includegraphics[width=\linewidth]{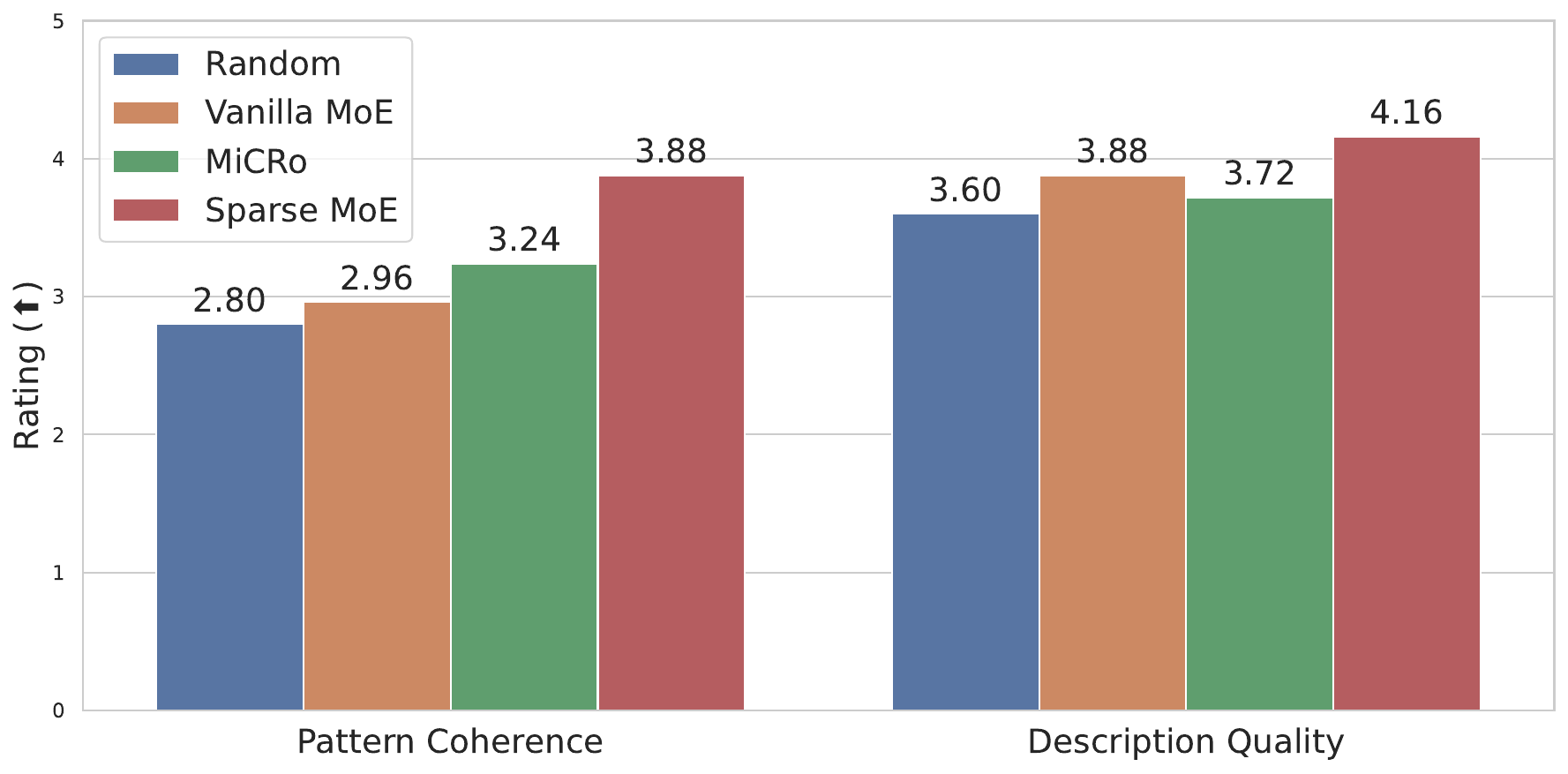}
    \caption{\textbf{Human evaluation of model interpretability.}
    Annotators find sparse MoE has the most interpretable routing patterns.
    The improvements in pattern coherence are statistically significant ($p < 0.05$).}
    \label{fig:human_study}
    \vspace{-5mm}
\end{wrapfigure}

\paragraph{Human Evaluation} 
We also check whether the routing patterns are coherent to humans.
For each model, we randomly sample five experts and present their ten highest-routed examples to five annotators, who rate \textit{pattern coherence} and \textit{description quality} on a 1--5 scale (Appendix~\ref{app:human_study}).
Sparse MoE receives the highest ratings on both dimensions, with the largest gain in pattern coherence (Figure~\ref{fig:human_study}), corroborating the automatic evaluation.

\vspace{-0.5em}
\begin{center}
\colorbox{gray!15}{%
  \parbox{0.98\linewidth}
{\textbf{Takeaway (coherence and correspondence at scale)}: On real-world preference data, sparse MoE discovers interpretable routing structure aligned with distinct expert strengths.}} 
\end{center}
 \vspace{-0.5em}

\subsection{Composing Interpretable Experts for Preference Adaptation}
\label{sec:adaptation}

We now turn to the third rung: whether the structure is composable as well as legible. We conduct test-time preference adaptation, freezing the experts and updating only the router on a small adaptation set.

\paragraph{Evaluation Setting}
Following~\citet{luo-etal-2025-rethinking}, we target the seven RPR attributes with the most examples, using $n=50$ training examples for adaptation and evaluating on all test examples.
We adapt the MoE models trained on 700K, which never see RPR's attribute annotations, freezing all experts and updating only the router with the Hedge algorithm~\citep{shen-etal-2025-micro}.
We also compare against HyRe~\citep{test-time-alignment-2024}, an ensemble of 20 reward models whose combination weights are adapted at test time, and ArmoRM~\citep{wang-etal-2024-interpretable}, an 8B MoE reward model whose experts are trained with attribute annotations.
We report post-adaptation accuracy and improvement over the corresponding unadapted model, averaged over three independently sampled adaptation sets.

\paragraph{Results}
Table~\ref{tab:rpr_personalization} shows that sparse MoE benefits substantially more from router adaptation than the competing methods.
With only 50 adaptation examples its overall accuracy rises by 25.81 points to 72.10\%, against 15.33 for SARM and roughly 9--10 for vanilla MoE and MiCRo, giving it both the largest gain and the highest post-adaptation accuracy.
Because vanilla MoE and MiCRo share sparse MoE's router and adaptation algorithm, this gap isolates the effect of expert structure rather than of the adaptation interface; HyRe, which adapts 20 ensemble weights with no expert specialization, gains only 5.98 points.
The gains are largest on attributes where the unadapted model performs poorly, indicating that the experts can be recombined to represent preferences far from the universal preference function.
Moreover, Appendix~\ref{app:inspecting preferences} shows that the sparse MoE experts that receive the largest weight increases after adaptation often exhibit plausible semantic associations with the target attributes, further linking effective personalization to expert interpretability.


\begin{table*}[ht]
    \centering
    \caption{\textbf{Personalization results on RPR.}
    Scores in parentheses denote changes after adaptation. 
    Sparse MoE shows the strongest overall personalization performance.
    Appendix~\ref{app:num_adaptation_set} further shows that a 14-point improvement is already reached with 5 examples.
    }
    \resizebox{\textwidth}{!}
    {
    \begin{tabular}{lcccccccc}
    \toprule
     Method & \makecell{Linguistic\\Creativity} & \makecell{User\\Friendliness} & \makecell{Humor\\ \& Entertainment} & \makecell{Scientific\\Rigor} & \makecell{Narrative \\ \& Storytelling} & \makecell{Creativity \\ \& Originality} & \makecell{Factual\\Accuracy}  & Overall \\
     \midrule
Single Reward & 31.73 & 39.33 & 30.16 & 76.19 & 34.17 & 18.52 & 85.92 & 45.15 \\
SARM & 62.82 (+19.55) & \textbf{60.30} (+13.11) & \textbf{83.73} (+50.40) & 62.70 (-3.97) & 60.83 (+10.83) & \textbf{61.11} (+31.94) & 57.28 (-14.55) & 64.11 (+15.33) \\
HyRe & 37.50 (+6.73) & 46.44 (+9.36) & 32.14 (+3.57) & 78.17 (+0.79) & 40.00 (+7.50) & 31.02 (+14.35) & 81.22 (-0.47) & 49.50 (+5.98) \\
Vanilla MoE & 42.31 (+11.54) & 44.19 (+8.23) & 36.51 (+6.75) & 81.74 (+3.17) & 51.67 (+19.17) & 28.70 (+10.64) & \textbf{89.20 (+7.51)} & 53.47 (+9.57) \\
MiCRo & 40.39 (+9.62) & 42.32 (+6.36) & 40.48 (+10.72) & 81.35 (+2.78) & 51.25 (+18.75) & 28.24 (+10.18) & 88.73 (+7.04) & 53.25 (+9.35) \\
ArmoRM & 21.15 (+0.00) & 42.32 (+2.99) & 14.29 (+0.00) & \textbf{84.52} (+0.00) & 32.50 (+0.00) & 23.61 (+1.39) & 81.69 (+0.00) & 42.87 (+0.63) \\
\midrule
Sparse MoE & \textbf{74.04 (+43.27)} & 55.06 \textbf{(+14.61)} & 80.56 \textbf{(+50.80)} & 82.54 \textbf{(+7.54)} & \textbf{68.75 (+28.75)} & 60.19 \textbf{(+31.02)} & 83.57 (+4.70) & \textbf{72.10 (+25.81)} \\
    \bottomrule
\end{tabular}
    }
    \label{tab:rpr_personalization}
    \vspace{-0.5em}
\end{table*}

The personalization improvement is also visible at the user level: on PersonalLLM~\citep{zollo2025personalllm}, sparse MoE gains 2.47 points after adaptation whereas vanilla MoE and MiCRo \emph{lose} accuracy ($-0.74$ and $-0.90$; Appendix~\ref{app:user_personalization}), so entangled components are not only harder to interpret but also harder to recombine.
Appendix~\ref{app:generalization} reports similar gains across backbones and datasets.
The adapted reward models can also serve as optimization signals for policy training: for two attributes, GRPO post-training against an adapted sparse MoE reward model yields responses preferred over those from a policy trained against the unadapted model (53.0 versus 45.5 win rate on creativity, 52.0 versus 44.5 on scientific rigor), without reducing overall quality (Appendix~\ref{app:post_train}).


\vspace{-0.5em}
\begin{center}
\colorbox{gray!15}{%
  \parbox{0.98\linewidth}{\textbf{Takeaway (composability)}: The disentangled, specialized experts sparse MoE learns are also composable: lightweight router adaptation recombines them for unseen preferences. The same adaptation gains far less on entangled experts, and degrades at the user level.}}
\end{center}

\section{Conclusion}

We investigated whether latent experts in MoE reward models can become both interpretable and useful for personalization.
Conventional training leaves experts entangled, whereas simple sparsity and diversity regularization yields coherent, functionally specialized experts whose roles are supported by targeted interventions.
The same specialization also makes experts more composable, leading to substantially stronger few-shot personalization across settings.
Although this comes with a modest reduction in universal preference accuracy, the resulting expert structure provides a more effective basis for adapting reward models to unseen preferences.
\footnote{Limitations of our work are discussed in Appendix~\ref{app:limitations}.}

\section*{Acknowledgements}
This work was funded by the DFG project GRK 2853 "Neuroexplicit Models of Language, Vision, and Action" (project number 471607914).

\section*{AI Use Statement}

In this work, we used generative AI tools for coding assistance, text polishing, related work discovery, and figure generation. 
We have not used generative AI tools for idea discovery or development of the conceptual framework.
All scientific content, experimental decisions, analyses, and conclusions were developed and verified by the authors.

\section*{Reproducibility Statement}

The code, data, configurations, and prompts used to implement and reproduce our experiments are described in the paper and included with the submission.
We will make the code base publicly accessible after the paper is accepted.

\section*{Ethics Statement}
While our work aims to address the challenge of modeling heterogeneous human values, the models we study also introduce potential risks of dual use, as they can be adapted to overfit to individual preferences.
As a result, universal principles such as fairness and safety constraints could potentially be bypassed, and harmful behaviors or addictive usage patterns may be reinforced.
In addition, making certain preference patterns more interpretable may increase the risk of privacy leakage or misuse of sensitive user information.
Therefore, we strongly recommend that any real-world deployment of such systems undergo careful ethical review and oversight by appropriate ethics committees.

\bibliography{iclr2027_conference}
\bibliographystyle{iclr2027_conference}

\appendix

\section{Details on Experimental Setup}
\label{app:experimental details}

\paragraph{SHP}
The SHP dataset is curated from Reddit posts and contains 18 subsets featuring different domains.
Following the recommended practice from the dataset curator, we use only examples where both chosen and rejected responses receive more than 3 likes.
For training and evaluation, we exclude the \textit{explainlikeimfive} and \textit{changemyview} subsets, as their examples contain special prefixes that can easily leak the domain information.
For examples replying to the same input, we only sample one pair of responses for evaluation.
In the end, we use 36000/4000/1497 examples for train/val/test sets, respectively.
The domains, their groupings, and numbers in the test set can be found in Table~\ref{tab:shp_details}.

\begin{table*}[ht]
    \centering
    \caption{\textbf{Domain composition and number of test examples for each category in SHP.}
    }
    \resizebox{\linewidth}{!}
    {
    \begin{tabular}{lcc}
    \toprule
     Category & Domain & \# Examples in the Test Set   \\ 
     \midrule
    Academic & askphysics, askscience, askengineers, askhistorians, askphilosophy, asksocialscience, askanthropology & 617 \\
    Professional & askdocs, askvet, askhr, legaladvice, askacademia, askcarguys & 384 \\
    Cooking & askbaking, askculinary & 256 \\
    Fiction & asksciencefiction & 240 \\
    
    \bottomrule
    \end{tabular}
    }
    
    \label{tab:shp_details}
\end{table*}

\paragraph{RPR}

Since each example in RPR is paired with two attributes, under which the preference is swapped, we convert each example to two attribute-conditioned examples with context during training.
Specifically, following~\citet{shen-etal-2025-micro}, we append the criteria corresponding to target attributes to the inputs and assign labels accordingly.
We use the complete training set with all 40 attributes for reward model training, resulting in 18300/2034 train/validation examples.
Some examples of attributes and their corresponding context-dependent criteria are provided below:

\begin{itemize}
    \item Linguistic Creativity: The response should demonstrate linguistic creativity by incorporating poetry or quotations attributed to the Sufi figure, enhancing the cultural and emotional understanding of their teachings.
    \item Detail and Elaboration: Provides a detailed step-by-step guide on how to build the app.
    \item Clarity and Conciseness: Provides a precise and succinct explanation of the bug, focusing on its technical aspects and why it violates design guidelines.
\end{itemize}

For evaluation robustness, we conduct attribute steering on the attributes that appear over 50 times in the test set, and personalization on the seven of these with the most examples.
The attributes appearing over 50 times are: \textit{Linguistic Creativity, User Friendliness, Humor and Entertainment Value, Scientific Rigor, Narrative and Storytelling Quality, Creativity and Originality, Factual Accuracy, Innovativeness, Pedagogical Effectiveness, Economic Feasibility, Technical Complexity, Interdisciplinary Approach, Empathy and Emotional Intelligence, Contextual Relevance, Clarity and Conciseness, User Experience, Practical Application}.

We perform paired $t$-tests on flip rates across all tested attributes to assess the statistical significance of the improvements.

\paragraph{700K}
We randomly sample 36000/4000/5000 examples to create the train/validation/test sets for 700K.

\paragraph{License}
We use publicly available datasets and models for research purposes and cite the original sources.
SHP is derived from Reddit data under the Reddit API Terms of Use and does not provide a standard open-source license.
RPR is released under CDLA-Permissive-2.0.
For preference-700K, we could not identify an explicit license on the dataset card; therefore, we report the license of the official accompanying RLHF Workflow repository, which is released under Apache-2.0.
For PersonalLLM, although the Hugging Face metadata lists CC BY 4.0, the paper states that the dataset is released under CC BY-NC 4.0; we therefore conservatively treat it as non-commercial research data.
The GRM-Llama3.2-3B reward model and Qwen3.6 models used for expert interpretation and evaluation are released under Apache-2.0.
We do not redistribute the original datasets or model weights.

\paragraph{Training Details}
The router is implemented as a one-hidden-layer MLP with 128 hidden units, and each expert is modeled as a single linear head on top of the backbone model representations.
For a fair comparison, we use the same training configuration for all models and baselines.
We set the learning rate to 0.002/0.0005 for experts and router parameters, respectively, with a default AdamW optimizer and 5\% warmup steps.
A dropout rate of 0.2 and jitter noise of 0.25 are applied to the router in MoE models to avoid routing collapse.

We train the models for 3 epochs with a batch size of 16 on one H100 GPU, as we find sufficiently long training is necessary for MoE to develop reasonable specialization. 

\paragraph{Evaluation Metrics}

We provide mathematical descriptions of the evaluation metrics we used in the experiments.

\par\vspace{0.4em}
\noindent \textbullet\ \textbf{Category Coverage} Let $\mathcal{C}$ denote the set of ground truth categories. For each expert $k$,
we assign a category label $\hat{c}_k$ based on the majority category
among its top-activating validation examples. Category coverage is defined as:
\begin{equation}
\mathrm{Coverage}
=
\frac{
\left|
\{c \in \mathcal{C}: \exists k, \hat{c}_k = c\}
\right|
}{
|\mathcal{C}|
}.
\end{equation}

\noindent \textbullet\ \textbf{Expert Purity} Let $\ell(x) \in \mathcal{C}$ denote the ground truth category of input
$x$. For each expert $k$, let $\mathcal{T}_k^{(n)}$ be the set
of $n$ test examples with the largest routing weights $\pi_{\phi,k}(x)$.
Given the category label $\hat{c}_k$ assigned to expert $k$, expert
purity is defined as:
\begin{equation}
\mathrm{Purity}_k
=
\frac{1}{|\mathcal{T}_k^{(n)}|}
\sum_{x \in \mathcal{T}_k^{(n)}}
\mathbf{1}\{\ell(x)=\hat{c}_k\}.
\end{equation}

The overall expert purity score is averaged across experts:
\begin{equation}
\mathrm{Purity}
=
\frac{1}{K}
\sum_{k=1}^{K}
\mathrm{Purity}_k.
\end{equation}

\noindent \textbullet\ \textbf{Expert Specialization}
Let $\mathcal{D}_c$ denote the set of test examples belonging to
category $c$, and let $\mathrm{Acc}_j(c)$ denote the accuracy of expert
$j$ on $\mathcal{D}_c$. For each expert $k$ with assigned category
$\hat{c}_k$, we define its specialization score as the difference
between its accuracy on $\hat{c}_k$ and the average accuracy of all
experts on the same category:
\begin{equation}
\mathrm{Spec}_k
=
\mathrm{Acc}_k(\hat{c}_k)
-
\frac{1}{K}
\sum_{j=1}^{K}
\mathrm{Acc}_j(\hat{c}_k).
\end{equation}

In settings without ground truth categories, we instead evaluate
specialization on the examples most strongly associated with each
expert. For each expert $k$, let $\mathcal{T}_k^{(m\%)}$ denote the
top m\% test examples with the largest routing weights
$\pi_{\phi,k}(x)$. Let $\mathrm{Acc}_j(\mathcal{S})$ denote the
accuracy of expert $j$ on a set of examples $\mathcal{S}$. We define
the specialization score of expert $k$ as:
\begin{equation}
\mathrm{Spec}_k
=
\mathrm{Acc}_k(\mathcal{T}_k^{(m\%)})
-
\frac{1}{K}
\sum_{j=1}^{K}
\mathrm{Acc}_j(\mathcal{T}_k^{(m\%)}).
\end{equation}

The overall expert specialization score is averaged across experts:
\begin{equation}
\mathrm{Spec}
=
\frac{1}{K}
\sum_{k=1}^{K}
\mathrm{Spec}_k .
\end{equation}

\noindent \textbullet\ \textbf{Desired Flip Rate}
Let $\mathcal{A}$ denote the set of evaluated attributes and
$\mathcal{W}$ denote the set of target expert weights. For each
attribute $a \in \mathcal{A}$, let $\mathcal{D}_a$ be its test set and
let $k_a$ be the target expert selected for this attribute. Given a
target weight $w \in \mathcal{W}$, we define the intervened routing
distribution $\tilde{\pi}_{\phi}^{(k_a,w)}(x)$ by setting the weight of
expert $k_a$ to $w$ and rescaling the remaining expert weights to sum
to $1-w$.

Let $\hat{y}^{(k_a,w)}(x)$ denote the response selected by the model
under this intervention, and let $y_a^+(x)$ denote the response
preferred under attribute $a$. We first identify the examples that are
misclassified when the target expert is removed:
\begin{equation}
\mathcal{M}_a
=
\left\{
x \in \mathcal{D}_a :
\hat{y}^{(k_a,0)}(x) \neq y_a^+(x)
\right\}.
\end{equation}

The desired flip rate for attribute $a$ under target expert weight $w$ is:
\begin{equation}
\mathrm{DFR}(a,w)
=
\frac{\sum_{x \in \mathcal{M}_a}
\mathbf{1}
\left\{
\hat{y}^{(k_a,w)}(x) = y_a^+(x)
\right\}}{|\mathcal{M}_a|}
.
\end{equation}

The overall desired flip rate is averaged over all evaluated attributes and
target expert weights:
\begin{equation}
\mathrm{DFR}
=
\frac{1}{|\mathcal{A}|\,|\mathcal{W}|}
\sum_{a \in \mathcal{A}}
\sum_{w \in \mathcal{W}}
\mathrm{DFR}(a,w).
\end{equation}

\noindent \textbullet\ \textbf{Undesired Flip Rate}
For each attribute $a$, we define the
set of initially correct examples under target expert removal as:
\begin{equation}
\mathcal{C}_a
=
\left\{
x \in \mathcal{D}_a :
\hat{y}^{(k_a,0)}(x) = y_a^+(x)
\right\}.
\end{equation}
The undesired flip rate under target weight $w$ is then:
\begin{equation}
\mathrm{UFR}(a,w)
=
\frac{\sum_{x \in \mathcal{C}_a}
\mathbf{1}
\left\{
\hat{y}^{(k_a,w)}(x) \neq y_a^+(x)
\right\}}{|\mathcal{C}_a|}
.
\end{equation}

\noindent \textbullet\ \textbf{Description Fidelity}
For each expert $k$, let $d_k$ denote its natural language description,
and let $\mathcal{S}_k$ be the set of sampled test examples used for
evaluation. For each example $x \in \mathcal{S}_k$, we ask an LLM to
judge whether $x$ matches $d_k$, producing a binary relevance judgment
$z_k(x) \in \{0,1\}$. We then compute the Pearson correlation between
these judgments and the corresponding router activations
$\pi_{\phi,k}(x)$:
\begin{equation}
\mathrm{DF}_k
=
\mathrm{corr}_{x \in \mathcal{S}_k}
\left(
z_k(x), \pi_{\phi,k}(x)
\right).
\end{equation}

The overall description fidelity score is averaged across experts:
\begin{equation}
\mathrm{DF}
=
\frac{1}{K}
\sum_{k=1}^{K}
\mathrm{DF}_k .
\end{equation}

\paragraph{Expert Interpretation}

When computing the description fidelity of expert interpretations, since only a small fraction of examples strongly activate any given expert, we stratify the sample by drawing 100 examples from the top 10\% of activations and 100 examples from the remaining 90\%.

The prompts that we use for interpreting experts and evaluating interpretability are adapted from~\citet{movva2026whats} and are provided in Figure~\ref{fig:prompt_describe_experts} and Figure~\ref{fig:prompt_match_description}.

\begin{figure*}[t]
\centering
\begin{promptbox}{Prompt: Describing experts in natural language given top-activating examples}
{\ttfamily\scriptsize

\textbf{Task Overview}

You are a machine learning researcher analyzing a sparse mixture-of-experts reward model.

The model contains a router that assigns each input prompt to one or more experts. You are trying to understand what feature of the input prompt causes a specific expert to receive high routing weight.

Each dataset example consists only of a user prompt. These prompts were selected because this expert was strongly activated for them, meaning the router assigned this expert a high weight.

Your goal is to identify the common feature shared by these prompts, and to describe this feature as a concise, natural language concept.

\medskip
\textbf{Instructions}

First, inspect each prompt and consider what concrete property it has.\\
Then, synthesize the single feature that is most consistently shared across the prompts.

The feature should describe the input prompt itself, not the model, the router, or the expert.

For example, the common feature might be:
\begin{itemize}[leftmargin=2em, itemsep=2pt, topsep=2pt]
    \item "asks for emotionally supportive advice"
    \item "requests factual explanation of a technical concept"
    \item "asks for creative writing or brainstorming"
    \item "contains a safety-sensitive or potentially harmful request"
    \item "asks for evaluation of competing options"
\end{itemize}

If the prompts do not share a clear topic, focus on the type of user intent, task format, reasoning style, domain, constraint, or interaction pattern that best explains why the expert fires.

\medskip
\textbf{Additional Rules}

Rules about the feature you identify:
\begin{itemize}[leftmargin=2em, itemsep=2pt, topsep=2pt]
    \item The feature should be objective and concrete.
    \item Err on the side of being concise.
    \item The feature should be an attribute that could be present in an individual prompt.
    \item Do not refer to "the expert", "the router", "activation", or "these examples" in the final answer.
    \item Do not use vague descriptions such as "is helpful", "is complex", or "is high quality".
    \item Avoid overly broad concepts such as "general question" unless no more specific feature is apparent.
    \item If the shared feature is a negation, describe it directly, e.g., "does not specify a concrete task or goal".
\end{itemize}

\medskip
\textbf{Prompts}

----------------\\
\{examples\}\\
----------------

\medskip
\textbf{Output}

Do not output anything besides the feature.\\
Output exactly one description, starting with "-" and surrounded by quotes.

\medskip
Your response is:-"
}
\end{promptbox}
\caption{\textbf{Prompt used to summarize the common feature of top-activating prompts and describe it in natural language.}}
\label{fig:prompt_describe_experts}
\end{figure*}

\begin{figure}[t]
\centering
\begin{promptbox}{Prompt: Judging whether an example matches expert description}
{\ttfamily\scriptsize

You are a binary classifier.

\medskip
\textbf{Task:}\\
Determine whether the input text expresses the feature described below.

\medskip
\textbf{Feature description:}\\
\{description\}

\medskip
\textbf{Input text:}\\
\{input\_text\}

\medskip
\textbf{Decision rule:}\\
Answer YES if the input text clearly matches or expresses the feature description.\\
Answer NO otherwise.

\medskip
\textbf{Output format:}\\
Return exactly one token: YES or NO.

\medskip
\textbf{Answer:}
}
\end{promptbox}
\caption{\textbf{Prompt used to judge whether an input example matches a given natural language feature description.}}
\label{fig:prompt_match_description}
\end{figure}

\paragraph{Test-Time Personalization}
We use the test-time weight adjustment method from~\citet{test-time-alignment-2024} for HyRe and implement the Hedge algorithm from~\citet{shen-etal-2025-micro} for fine-tuning routers in MoE models.
For Hedge, we use $\tau = 0.001$, following the recommendation of the original paper.
All models are trained for 10 epochs using a batch size of 32 and a learning rate of 0.01 for adaptation.

\section{Effect of the Number of Experts}
\label{app:num_expert}

We study the effect of the number of experts by training sparse MoE models with
$K \in \{10, 20, 30, 40, 50\}$ on 700K and evaluating their interpretability.
We conduct this analysis on the real-world dataset rather than the controlled settings, because the pre-defined labels in the controlled experiments capture only coarse latent structures and may not fully reflect the finer-grained patterns discovered by the model.

\begin{wrapfigure}{r}{0.48\linewidth}
    \centering
    \vspace{-5mm}
    \includegraphics[width=1.0\linewidth]{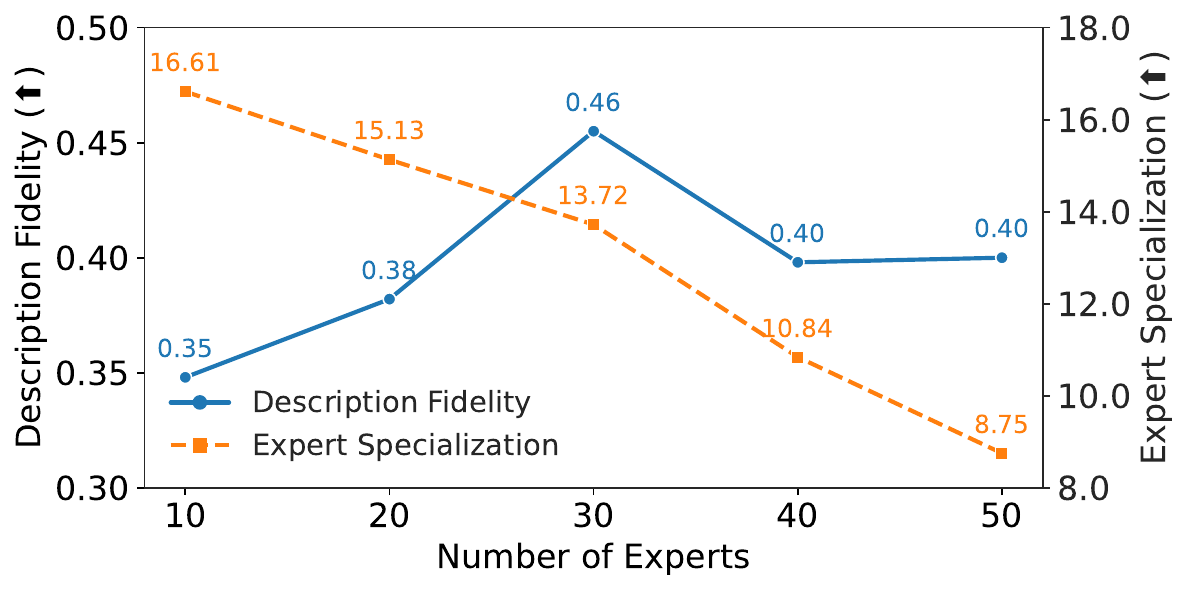}
    \caption{\textbf{Effect of the number of experts on interpretability.}
    Description fidelity peaks at a certain number of experts, while expert specialization consistently decreases as the number of experts increases.
    }
    \label{fig:expert_number}
    \vspace{-3mm}
\end{wrapfigure}

Figure~\ref{fig:expert_number} shows how description fidelity and expert specialization change with the number of experts (single run with seed=42).
Description fidelity increases from $K=10$ to $K=30$, where it reaches its peak, and then remains relatively stable.
This suggests that increasing the number of experts initially helps the router discover more semantically coherent input clusters, but adding experts beyond this point provides limited additional benefit.
We interpret this peak as reflecting the effective granularity of input patterns that can be reliably supported by the data, training objective, and interpretation procedure.

By contrast, expert specialization decreases steadily as $K$ increases.
This suggests that larger expert counts may fragment the preference learning signal and weaken the behavioral specialization of individual expert heads.
Together, these trends indicate a trade-off between semantic interpretability and functional specialization: router assignments can remain semantically interpretable at larger $K$, while the corresponding expert heads become less behaviorally specialized.
We therefore recommend selecting $K$ based on the complexity of the target data distribution.

\section{Ablation Study of Interpretability Regularization}
\label{app:ablation}
To better understand the effect of each interpretability regularization term on model behavior, we conduct a comprehensive ablation study using a grid search over $\lambda_{ls} \in [0.0, 0.5, 1.0]$, $\lambda_{gb} \in [0.0, 0.5, 1.0]$, and $\lambda_{div} \in [0.0, 1.0, 2.0]$ on SHP with $K=5$ experts. The resulting task performance, routing pattern, expert behavior, and interpretability metrics are reported in Table~\ref{tab:ablation} (single run with seed=42).

For ease of visualization, we highlight in \textbf{bold} the models that do not suffer from collapse. 
These models exhibit low local entropy and high global entropy, indicating that the MoE does not rely on a single expert nor distribute inputs uniformly across experts. 
We observe that the interaction between $\lambda_{ls}$ and $\lambda_{gb}$ largely determines whether the trained MoE model collapses.
These two hyperparameters influence routing patterns in the intended manner: 
when the model uniformly utilizes all experts (i.e., high local and global entropy), increasing $\lambda_{ls}$ encourages more confident routing decisions; 
conversely, when the model collapses to a single expert (i.e., low local and global entropy), increasing $\lambda_{gb}$ encourages the model to utilize a broader set of experts.

While $\lambda_{ls}$ and $\lambda_{gb}$ primarily control router behavior, $\lambda_{div}$ directly affects model interpretability and expert specialization. 
Increasing $\lambda_{div}$ encourages more diverse expert outputs, leading to stronger expert specialization. 
At the same time, it imposes additional pressure on the router to distribute examples in a more semantically meaningful manner, resulting in higher expert purity.

Overall, although the configuration we study involves three regularization terms and their corresponding hyperparameters, the ablation study shows that they serve distinct roles, allowing users to tune them based on observed model behavior.
Experiments in Appendix~\ref{app:generalization} further demonstrate the robustness of our hyperparameter choices across different backbone models and datasets.
When training sparse MoE reward models on new datasets, we therefore recommend starting from the hyperparameter configuration that works well on known data, then tuning $\lambda_{ls}$ and $\lambda_{gb}$ to ensure stable training. 
$\lambda_{div}$ can subsequently be adjusted according to the desired degree of interpretability.

\begin{table*}
    \centering
    \caption{\textbf{Ablation study of interpretability regularization terms on SHP.} 
    Models that do not suffer from collapse are marked in \textbf{bold}.
    $\lambda_{ls}$ and $\lambda_{gb}$ mainly determine whether the MoE model collapses, while $\lambda_{div}$ affects model interpretability and expert specialization.
    }
    \resizebox{\textwidth}{!}{
    \begin{tabular}{ccc|cccc|cc}
    \toprule
    $\lambda_{ls}$ & $\lambda_{gb}$ & $\lambda_{div}$ & Accuracy ($\uparrow$) & Routing Entropy ($\downarrow$) & Global Entropy ($\uparrow$) & Expert Correlation ($\to 0$) & Expert Purity ($\uparrow$) & Expert Specialization ($\uparrow$) \\
    \midrule
 0.00 & 0.00 & 0.00 & 70.67 & 0.991 & 0.993 & 0.556 & 74.40 & -1.00 \\
 0.00 & 0.00 & 1.00 & 70.70 & 0.133 & 0.139 & 0.024 & 80.00 & 1.12 \\
 0.00 & 0.00 & 2.00 & 71.10 & 0.014 & 0.015 & 0.003 & 70.00 & 0.47 \\
 0.00 & 0.50 & 0.00 & 71.17 & 1.000 & 1.000 & 0.493 & 43.20 & -0.27 \\
 0.00 & 0.50 & 1.00 & 71.54 & 1.000 & 1.000 & 0.141 & 46.40 & 0.62 \\
 0.00 & 0.50 & 2.00 & 71.49 & 1.000 & 1.000 & 0.126 & 49.20 & -0.07 \\
 0.00 & 1.00 & 0.00 & 71.24 & 1.000 & 1.000 & 0.520 & 56.80 & -0.20 \\
 0.00 & 1.00 & 1.00 & 70.79 & 1.000 & 1.000 & 0.153 & 47.60 & -0.91 \\
 0.00 & 1.00 & 2.00 & 71.92 & 1.000 & 1.000 & 0.110 & 47.20 & 0.84 \\
 0.50 & 0.00 & 0.00 & 70.24 & 0.000 & 0.000 & 0.743 & 54.80 & -0.57 \\
 0.50 & 0.00 & 1.00 & 71.44 & 0.000 & 0.000 & -0.028 & 46.80 & -0.71 \\
 0.50 & 0.00 & 2.00 & 71.19 & 0.000 & 0.000 & -0.027 & 44.00 & -0.93 \\
\textbf{0.50} & \textbf{0.50} & \textbf{0.00} & \textbf{71.39} & \textbf{0.087} & \textbf{0.984} & \textbf{0.782} & \textbf{43.20} & \textbf{0.06} \\
\textbf{0.50} & \textbf{0.50} & \textbf{1.00} & \textbf{68.55} & \textbf{0.131} & \textbf{0.983} & \textbf{0.016} & \textbf{75.20} & \textbf{4.66} \\
\textbf{0.50} & \textbf{0.50} & \textbf{2.00} & \textbf{69.62} & \textbf{0.200} & \textbf{0.986} & \textbf{0.011} & \textbf{85.60} & \textbf{5.77} \\
\textbf{0.50} & \textbf{1.00} & \textbf{0.00} & \textbf{71.40} & \textbf{0.084} & \textbf{0.989} & \textbf{0.746} & \textbf{46.80} & \textbf{0.34} \\
\textbf{0.50} & \textbf{1.00} & \textbf{1.00} & \textbf{66.96} & \textbf{0.293} & \textbf{0.960} & \textbf{0.067} & \textbf{54.00} & \textbf{4.12} \\
\textbf{0.50} & \textbf{1.00} & \textbf{2.00} & \textbf{69.45} & \textbf{0.333} & \textbf{0.917} & \textbf{0.033} & \textbf{67.20} & \textbf{3.34} \\
 1.00 & 0.00 & 0.00 & 70.52 & 0.000 & 0.000 & 0.673 & 46.40 & -0.69 \\
 1.00 & 0.00 & 1.00 & 70.64 & 0.000 & 0.000 & -0.026 & 54.00 & -1.23 \\
 1.00 & 0.00 & 2.00 & 70.76 & 0.000 & 0.000 & 0.006 & 34.00 & -0.76 \\
\textbf{1.00} & \textbf{0.50} & \textbf{0.00} & \textbf{70.31} & \textbf{0.214} & \textbf{0.909} & \textbf{0.783} & \textbf{49.60} & \textbf{-0.06} \\
 1.00 & 0.50 & 1.00 & 70.05 & 0.142 & 0.148 & 0.044 & 80.40 & 1.10 \\
 1.00 & 0.50 & 2.00 & 70.54 & 0.161 & 0.166 & 0.007 & 92.00 & 1.35 \\
\textbf{1.00} & \textbf{1.00} & \textbf{0.00} & \textbf{71.17} & \textbf{0.083} & \textbf{0.983} & \textbf{0.746} & \textbf{47.20} & \textbf{0.38} \\
\textbf{1.00} & \textbf{1.00} & \textbf{1.00} & \textbf{69.31} & \textbf{0.088} & \textbf{0.987} & \textbf{0.072} & \textbf{47.20} & \textbf{3.71} \\
\textbf{1.00} & \textbf{1.00} & \textbf{2.00} & \textbf{66.15} & \textbf{0.104} & \textbf{0.997} & \textbf{0.009} & \textbf{54.80} & \textbf{3.98} \\
 \bottomrule
    \end{tabular}
    }
    
    \label{tab:ablation}
\end{table*}

\section{Generalization Results}
\label{app:generalization}

We train MoE models under two additional settings: using a different pretrained reward model with a distinct architecture, Qwen3-8B (Skywork-Reward-V2-Qwen3-8B), on the 700K dataset, and using the same Llama-3.2-3B model on a different preference dataset, LMArena.
We intentionally use the same hyperparameters across settings to evaluate the transferability of our hyperparameter choices.

Table~\ref{tab:generalization_interpretability} reports the interpretability of these models, compared with other methods implemented using the same respective base models and datasets.
Consistent with the main experiments, sparse MoE exhibits the strongest interpretability in both settings, achieving significantly higher description fidelity and expert specialization. 
These results support the generalizability of sparse MoE's interpretability across different base models and datasets, as well as the robustness of our hyperparameter selection.

Tables~\ref{tab:qwen_complete_rpr_personalization} and~\ref{tab:lmarena_personalization} further demonstrate that sparse MoE maintains strong personalization performance across base models and datasets. 
With the Qwen backbone, it substantially outperforms the baselines in both post-adaptation performance and personalization gains. 
When trained on LMArena, sparse MoE still achieves the largest personalization improvements. 
These results suggest that our core finding---that interpretable and specialized experts facilitate effective personalization---generalizes across settings.

    

    

\begin{table*}[t]

    \centering
    \caption{\textbf{Interpretability results across different backbone models and datasets.} Best scores are marked in \textbf{bold}.}

    \begin{subtable}[t]{0.49\textwidth}
        \centering
        \caption{700K with Qwen3-8B as backbone.}

        \resizebox{\linewidth}{!}{
        \begin{tabular}{lccc}
        \toprule
        Method 
        & Accuracy ($\uparrow$) 
        & Description Fidelity ($\uparrow$) 
        & Expert Specialization ($\uparrow$) \\
        \midrule
        Single Reward & 82.02 & - & - \\
        Vanilla MoE & 82.56 & 0.23 & 1.55 \\
        MiCRo & \textbf{83.02} & 0.01 & -0.09 \\
        Random MoE & - & -0.00 & -0.33 \\
        \midrule
        Sparse MoE & 82.64 & \textbf{0.41} & \textbf{16.75} \\
        \bottomrule
        \end{tabular}
        }
        \label{tab:qwen interpretability}
    \end{subtable}
    \hfill
    \begin{subtable}[t]{0.49\textwidth}
        \centering
        \caption{LMArena with Llama3.2-3B as backbone.}
        \resizebox{\linewidth}{!}{
        \begin{tabular}{lccc}
        \toprule
        Method 
        & Accuracy ($\uparrow$) 
        & Description Fidelity ($\uparrow$) 
        & Expert Specialization ($\uparrow$) \\
        \midrule
        Single Reward & \textbf{64.56} & - & - \\
        Vanilla MoE & 61.10 & 0.08 & 0.73 \\
        MiCRo & 60.18 & 0.07 & 0.91 \\
        Random MoE & - & -0.01 & 0.63 \\
        \midrule
        Sparse MoE & 60.22 & \textbf{0.23} & \textbf{7.88} \\
        \bottomrule
        \end{tabular}
        }
        \label{tab:lmarena interpretability}
    \end{subtable}
    
    \label{tab:generalization_interpretability}
\end{table*}

\begin{table*}[t]
    \centering
    \caption{\textbf{RPR personalization results with Qwen3-8B-based reward models trained on 700K.} Best scores are marked in \textbf{bold}.}
    \resizebox{\textwidth}{!}{
    \begin{tabular}{lcccccccc}
    \toprule
     Method & \makecell{Linguistic\\Creativity} & \makecell{User\\Friendliness} & \makecell{Humor\\ \& Entertainment} & \makecell{Scientific\\Rigor} & \makecell{Narrative \\ \& Storytelling} & \makecell{Creativity \\ \& Originality} & \makecell{Factual\\Accuracy}  & Overall \\
    \midrule

    Single Reward
    & 20.19
    & 31.46
    & 27.38
    & 83.33
    & 33.75
    & 26.39
    & 83.10
    & 43.66 \\

    HyRe
    & 25.64 (+7.37)
    & 30.71 (+0.37)
    & 30.16 (+6.35)
    & \textbf{90.48} \textbf{(+1.19)}
    & 37.92 (+4.17)
    & 35.65 (+6.48)
    & 84.51 (+0.00)
    & 47.87 (+3.70) \\

    Vanilla MoE
    & 53.20 (+33.01)
    & 49.06 (+15.35)
    & 56.75 (+31.75)
    & 82.14 (-3.57)
    & 57.92 (+26.67)
    & 72.68 (+46.29)
    & \textbf{89.67} (+0.94)
    & 65.92 (+21.49) \\

    MiCRo
    & 22.44 (+3.21)
    & 39.70 (+8.24)
    & 25.40 (+1.59)
    & 88.10 (-2.38)
    & 30.83 (+2.08)
    & 34.72 (+8.33)
    & 85.45 (-0.47)
    & 46.66 (+2.94) \\

    Sparse MoE
    & \textbf{83.97} \textbf{(+56.09)}
    & \textbf{61.05} \textbf{(+29.59)}
    & \textbf{97.22} \textbf{(+72.22)}
    & 85.71 \textbf{(+1.19)}
    & \textbf{67.50} \textbf{(+30.00)}
    & \textbf{81.02} \textbf{(+57.41)}
    & 84.98 \textbf{(+6.11)}
    & \textbf{80.21} \textbf{(+36.09)} \\

    \bottomrule
    \end{tabular}
    }
    \label{tab:qwen_complete_rpr_personalization}
\end{table*}

\begin{table*}[t]
    \centering
    \caption{\textbf{RPR personalization results with Llama3.2-3B-based reward models trained on LMArena.} Best scores are marked in \textbf{bold}.}
    \resizebox{\textwidth}{!}{
    \begin{tabular}{lcccccccc}
    \toprule
     Method & \makecell{Linguistic\\Creativity} & \makecell{User\\Friendliness} & \makecell{Humor\\ \& Entertainment} & \makecell{Scientific\\Rigor} & \makecell{Narrative \\ \& Storytelling} & \makecell{Creativity \\ \& Originality} & \makecell{Factual\\Accuracy}  & Overall \\
    \midrule

    Single Reward
    & 83.65
    & 49.44
    & 89.29
    & 28.57
    & 85.00
    & 79.17
    & 25.35
    & 62.92 \\

    HyRe
    & 86.10 (+12.06)
    & 71.16 \textbf{(+14.98)}
    & 90.25 (+0.96)
    & \textbf{69.22} \textbf{(+28.74)}
    & 80.33 (+5.33)
    & 82.65 (+15.98)
    & 80.10 (+40.66)
    & \textbf{79.97} (+16.96) \\

    Vanilla MoE
    & 88.46 (+14.42)
    & 60.67 (+2.24)
    & \textbf{93.26} \textbf{(+7.55)}
    & 61.11 (+23.01)
    & 86.25 (+13.75)
    & 86.57 (+21.29)
    & 74.65 (+32.40)
    & 78.71 (+16.38) \\

    MiCRo
    & 87.82 (+18.59)
    & 65.54 (+5.99)
    & \textbf{93.26} (+6.36)
    & 49.60 (+11.50)
    & \textbf{87.50} (+16.25)
    & \textbf{87.50} \textbf{(+23.61)}
    & 63.38 (+23.94)
    & 76.37 (+15.18) \\

    Sparse MoE
    & \textbf{89.42} \textbf{(+25.00)}
    & \textbf{72.28} (+9.36)
    & 88.89 (+6.75)
    & 63.89 (+26.99)
    & 77.08 \textbf{(+20.83)}
    & 81.02 (+22.69)
    & \textbf{81.69} \textbf{(+40.84)}
    & 79.18 \textbf{(+21.78)} \\

    \bottomrule
    \end{tabular}
    }
    \label{tab:lmarena_personalization}
\end{table*}

\section{Comparing MoE and SAE Reward Models}
\label{app:moe vs. sae}

In addition to the quantitative results, we also provide a qualitative comparison of MoE- and SAE-based interpretations.
In Table~\ref{tab:moe_vs_sae}, we sample five representative features or experts from two SAE-based methods, SARM and SparseRM~\citep{sparserm}, and compare them against sparse MoE interpretations.

We observe that SAE-based models learn qualitatively different types of features from sparse MoE. 
This difference may be explained by their training paradigms. 
Both SparseRM and SARM rely on SAE encoders pretrained on large-scale corpora, and then learn reward weights over the extracted features on 700K. Since SparseRM uses a token-level SAE, its features tend to capture lexical, formatting, or local syntactic patterns. 
SARM operates at the sequence level and therefore captures more response-level patterns, some of which are preference-relevant. 
However, despite their computationally expensive pretraining, both methods may suffer from a mismatch between SAE pretraining and the downstream preference-modeling objective.
As a result, their interpretations are less specific to the target dataset and yield worse quantitative interpretability results.

\begin{table*}[t]
    \centering
    \caption{\textbf{Representative feature/expert interpretations in sparse MoE and SAE-based reward models.} These examples reveal substantial qualitative differences in the types of patterns captured by the two model families.}
    \small
    \renewcommand{\arraystretch}{1.1}
    \setlength{\tabcolsep}{5pt}

    \begin{tabularx}{\textwidth}{YYY}
    \toprule
    \textbf{Sparse MoE} & \textbf{SARM} & \textbf{SparseRM} \\
    \midrule

    asks for advice, feedback, or discussion on a personal, professional, or academic challenge
    &
    provides a brief, direct, or conversational response to a user prompt
    &
    textual structures, particularly punctuation and formatting indicators
    \\
    \midrule

    asks to solve a procedural task involving numerical or list manipulation
    &
    provides a structured, step-by-step explanation or guide
    &
    specific verbs or actions associated with legal or formal contexts
    \\
    \midrule

    asks for a summary or extraction of information from a provided text
    &
    provides a direct, definitive answer to a factual or technical question
    &
    patterns and structures in data or questions
    \\
    \midrule

    asks for a direct answer to a multiple-choice or short-answer question
    &
    provides an incorrect, irrelevant, or hallucinated response to the prompt
    &
    specific names, titles, or identifiers related to individuals and organizations
    \\
    \midrule

    requests assistance with illegal, harmful, or sexually explicit activities
    &
    provides a step-by-step explanation or structured breakdown of the answer
    &
    elements related to programming or coding structures
    \\
    \bottomrule
    \end{tabularx}

    \label{tab:moe_vs_sae}
\end{table*}

\section{Full Results}
\label{app:complete_results}
\paragraph{Category Recovery Results}

Table~\ref{tab:complete_shp_results} and Table~\ref{tab:complete_shp_expert_results} present the overall performance of our sparse MoE model on SHP (single run with seed 42), together with the accuracy and routing weights of each expert.
We observe that the global entropy within each category is lower than that on the full dataset, indicating semantically reasonable routing patterns at the category level.
Examining the expert distributions further, we find that most categories assign more than 50\% of their routing weight to a single expert.
Moreover, these dominant experts typically achieve strong performance within their corresponding categories.
Together, these results suggest that the sparse MoE reward model learns meaningful routing patterns and behaviorally specialized experts.

We further analyze the performance drop of the sparse MoE reward model compared to the single-head baseline.
The largest degradation occurs in the academic and professional categories, where the best-performing experts do not receive the highest routing weights.
We therefore attribute the accuracy drop of sparse MoE mainly to a mismatch between routing weights and expert abilities.
This mismatch likely arises because these two categories are broad and heterogeneous, making it more difficult for the router to assign examples to the most suitable experts.

\begin{table*}[ht]
\centering
\caption{\textbf{Category-level evaluation results on SHP.}
Scores in parentheses denote performance comparison against the single-head reward model.
}
\resizebox{0.8\textwidth}{!}{
\begin{tabular}{lccccc}
\toprule
Domain & \# Examples & Accuracy ($\uparrow$) & Routing Entropy ($\downarrow$) & Global Entropy ($\uparrow$) & Expert Correlation ($\to 0$) \\
\midrule
Academic        & 617  & 70.50 (-7.06) & 0.252 & 0.883 &  0.079 \\
Professional    & 384  & 70.31 (-4.43) & 0.149 & 0.733 &  0.038 \\
Cooking            & 256  & 66.41 (-0.39) & 0.156 & 0.568 & -0.006 \\
Fiction & 240  & 71.25 (+0.84) & 0.198 & 0.704 &  0.046 \\
\midrule
Overall & 1497 & 69.62 (-0.90) & 0.200 & 0.986 &  0.011 \\
\bottomrule
\end{tabular}
}

\label{tab:complete_shp_results}
\end{table*}

\begin{table*}[ht]
\centering

\caption{\textbf{Expert accuracy and average routing weights on SHP.}
Each cell reports expert accuracy / routing weight. 
}
\resizebox{0.9\textwidth}{!}{
\begin{tabular}{lccccc}
\toprule
Domain & Expert 0 & Expert 1 & Expert 2 & Expert 3 & Expert 4 \\
\midrule
Academic 
& 65.80 / \textbf{35.56}
& 67.59 / 25.93
& 62.40 / 10.46
& 55.92 / 3.76
& \textbf{69.53} / 24.28 \\

Professional
& 58.85 / 10.98
& 52.34 / 6.91
& 67.45 / \textbf{55.40}
& 52.34 / 2.06
& \textbf{68.75} / 24.66 \\

Cooking
& 61.33 / 7.18
& 61.72 / 21.17
& 49.61 / 1.19
& \textbf{68.36} / \textbf{68.28}
& 59.77 / 2.18 \\

Fiction
& 63.33 / 15.81
& 56.25 / 4.33
& 66.25 / 28.57
& 46.25 / 0.08
& \textbf{68.33} / \textbf{51.21} \\

\midrule
Overall
& 62.86 / 21.23
& 60.86 / 16.78
& 62.12 / 23.31
& 55.58 / 13.77
& \textbf{67.47} / \textbf{24.92} \\
\bottomrule
\end{tabular}
}

\label{tab:complete_shp_expert_results}
\end{table*}

\paragraph{Expert Descriptions}
\begin{table*}[!t]
    \centering
        \caption{\textbf{Natural language descriptions of all experts in the sparse MoE model.}}
    \resizebox{\linewidth}{!}
    {
    \begin{tabular}{lccc}
    \toprule
     ID & Natural Language Description & Description Fidelity ($\uparrow$) & Expert Specialization ($\uparrow$)  \\ 
     \midrule
0 & asks for advice, feedback, or discussion on a personal, professional, or academic challenge & 0.43 &  22.08\\
1 & asks for natural language inference or textual entailment classification & 0.26 &  22.48\\
2 & requests the generation of computer code or technical scripts & 0.45 &  21.53\\
3 & asks for troubleshooting advice regarding a specific problem in cooking, baking, or home maintenance & 0.32 &  15.84\\
4 & asks for personal advice or recommendations & 0.50 &  6.80\\
5 & contains a request for harmful, illegal, or policy-violating content & 0.43 &  7.97\\
6 & asks to solve a procedural task involving numerical or list manipulation & 0.51 &  25.98\\
7 & asks for an opinion, consensus, or general advice on a subjective or non-factual topic & 0.41 &  6.20\\
8 & asks for a summary or extraction of information from a provided text & 0.45 &  19.35\\
9 & asks for general information, advice, or explanation on a broad topic & 0.62 &  8.59\\
10 & asks for a direct answer to a multiple-choice or short-answer question & 0.36 &  20.56\\
11 & asks for technical advice, architectural design, or problem-solving in software engineering and IT & -0.19 &  19.49\\
12 & requests technical implementation details, system design, or professional strategy plans & 0.39 &  18.69\\
13 & requests assistance with illegal, harmful, or sexually explicit activities & 0.62 &  3.35\\
14 & asks for legal or procedural advice regarding a specific real-world situation & 0.33 &  24.32\\
15 & asks to solve a competitive mathematics problem & 0.42 &  32.00\\
16 & asks for a list of examples, suggestions, or items & 0.30 &  8.60\\
17 & asks a question containing a factual error, category mistake, or logical impossibility & 0.30 &  3.92\\
18 & requests assistance with illegal, harmful, or unethical activities & 0.36 &  6.76\\
19 & asks a question containing a factual error, logical fallacy, or absurd premise & 0.35 &  8.12\\
    \bottomrule
    \end{tabular}
    }

    \label{tab:all_interpretation}
\end{table*}

Table~\ref{tab:all_interpretation} presents the natural language descriptions of all experts in the sparse MoE model trained on 700K examples.
Most experts specialize in a concrete type of question or task, exhibiting high description fidelity while also specializing behaviorally in the examples routed to them.
Despite nuanced differences between experts, many focus on recurring categories such as feedback-seeking, safety-related queries, mathematics, code generation, and question answering.
These expert patterns partially reflect the underlying structure and distribution of the training data.

We also observe that experts with similar activation patterns, measured by high correlations in their activations across examples, tend to receive similar interpretations. 
For instance, expert 4 exhibits strong activation correlations with expert 7 (0.95) and expert 16 (0.93), and all three are identified as advice-seeking experts. 
Similarly, safety-related experts also display highly aligned activation patterns, as illustrated by the 0.96 correlation between experts 5 and 18.

These findings suggest the potential for hierarchical expert interpretation and for reducing expert redundancy through additional regularization on router activation correlations, which we leave for future work.

\section{Details on Human Study}
\label{app:human_study}
We recruited five annotators with at least a master's degree in computer science, computational linguistics, or linguistics to evaluate the interpretability of sparse MoE and the baseline models.
Annotators were first asked to rate pattern coherence after seeing the top-activating examples for each expert.
They were then shown the LLM-generated natural language description and asked to rate its quality.

The ratings show fair to moderate inter-annotator agreement.
Sparse MoE achieves statistically significant improvements in pattern coherence over all baselines ($p < 0.05$), while the improvements in description quality are not statistically significant.
Statistical significance is computed with a paired $t$-test.

The annotation instructions and rating scales for pattern coherence and description quality are provided below.

\paragraph{Pattern Coherence} Pattern coherence measures whether the input prompts in the same group share a consistent and recognizable pattern. The shared pattern could be a specific task, shared topic, semantic feature, input format, writing style, user intent, domain, or another interpretable property. A high score means that the prompts appear meaningfully related and could plausibly be summarized by a common feature. A low score means that the prompts appear random, unrelated, or too diverse to be summarized by a single coherent pattern.

\begin{enumerate}
    \item \textbf{Not Coherent:} The prompts appear unrelated, random, or inconsistent. No clear shared task, topic, feature, format, style, intent, or domain can be identified.
    \item \textbf{Weakly Coherent:} A few prompts share some similarities, but the overall group is noisy and difficult to summarize with a single pattern.
    \item \textbf{Moderately Coherent:} The prompts share a partially recognizable pattern, but there are also several exceptions or mixed themes.
    \item \textbf{Mostly Coherent:} Most prompts share a clear common task, topic, feature, format, style, intent, domain, or other interpretable property, with only minor exceptions.
    \item \textbf{Highly Coherent:} The prompts are strongly and consistently related. They clearly reflect a common interpretable concept.
\end{enumerate}

\paragraph{Description Quality} 
Description quality measures how accurately the provided natural language description captures the common pattern shared by the input prompts. This pattern may involve a specific task, shared topic, semantic feature, input format, writing style, user intent, domain, or another interpretable property. A high score means that the description correctly summarizes the main shared pattern among the prompts without adding misleading or unsupported claims. A low score means that the description is inaccurate, too vague, inconsistent with the prompts, or misses the main pattern.

\begin{enumerate}
    \item \textbf{Poor:} The description does not match the prompts, misses the main pattern, or describes a feature that is mostly absent.
    \item \textbf{Limited:} The description captures only a small part of the pattern, or is mostly vague, incomplete, or partially incorrect.
    \item \textbf{Adequate:} The description captures some important shared features, but it is still vague, misses other key aspects or includes noticeable inaccuracies.
    \item \textbf{Good:} The description accurately captures the main pattern of the prompts, with only minor omissions or minor overgeneralizations.
    \item \textbf{Excellent:} The description clearly and accurately summarizes the shared features of the prompts, with no meaningful inaccuracies.
\end{enumerate}

\paragraph{Human Study Ethics and Compensation}
Before participating in the study, annotators were informed about how their data would be used and were told that their annotations would be deleted upon request.
The study took approximately 20 minutes on average, and annotators were compensated at an hourly rate comparable to the local average hourly salary for researchers.
The human study protocol has been approved by the institutional ethics board.

\begin{figure}[h]
    \centering
    \includegraphics[width=0.65\linewidth]{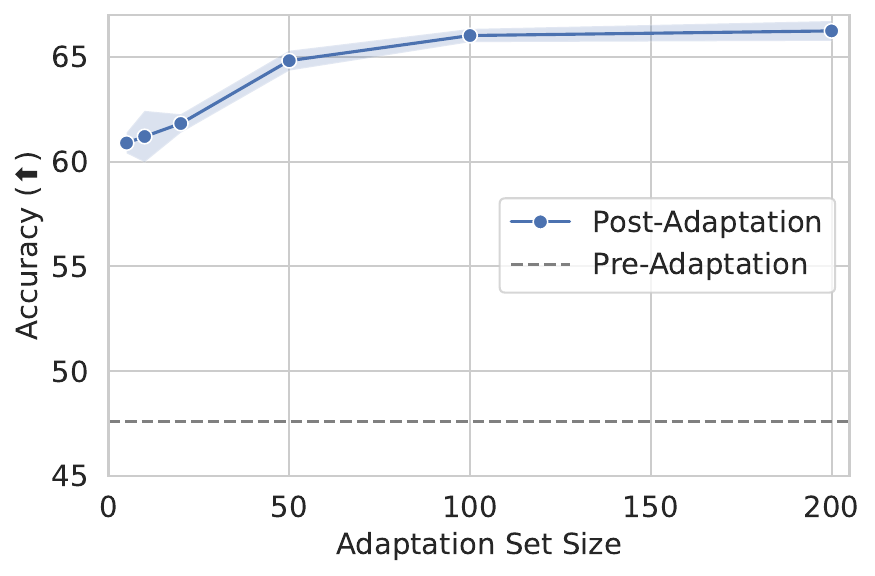}
    \caption{\textbf{Personalization results of sparse MoE with different adaptation set sizes on RPR.}
    Even with only 5 adaptation examples, sparse MoE achieves strong personalization performance.
    Performance further improves with larger adaptation sets before saturating at around 100 examples.
    }
    \label{fig:n_adaptation}
\end{figure}

\section{Effect of Adaptation Set Size}
\label{app:num_adaptation_set}

Figure~\ref{fig:n_adaptation} shows the attribute-level personalization performance of the sparse MoE model under different adaptation set sizes.
Results are averaged over the seven attributes and three independently sampled adaptation sets.
We observe that, even with only 5 adaptation examples, the model improves accuracy by approximately 14 points.
As expected, adaptation performance further improves with larger adaptation sets and begins to saturate at around 100 examples.

\section{Inspecting Preference Adaptation}
\label{app:inspecting preferences}

Leveraging the interpretability and specialization of experts in sparse MoE, we further analyze the model's adaptation to individual preferences by examining changes in expert weights after adaptation.
Table~\ref{tab:rpr_personalization_interpretation} reports, for a subset of attributes, the expert whose weight increases the most after adaptation.
We observe that the upweighted experts often show plausible semantic associations with the target attributes.
Overall, post-adaptation expert weight shifts provide a useful diagnostic signal for inspecting preference adaptation.

\begin{table}[t]
\caption{\textbf{Experts most strongly upweighted after adaptation for selected attributes.}
The experts align semantically with the adapted attributes in an interpretable way.}
\centering
\resizebox{0.8\linewidth}{!}{
\begin{tabular}{l p{0.7\linewidth}}
\toprule
Attribute  &  Most Upweighted Expert \\
\midrule
\makecell[tl]{Factual\\Accuracy}  & asks for a summary or extraction of information from a provided text \\
\midrule
\makecell[tl]{Pedagogical\\Effectiveness} & asks for advice, feedback, or discussion on a personal, professional, or academic challenge \\
\midrule
\makecell[tl]{Innovativeness} & asks a question containing a factual error, logical fallacy, or absurd premise \\
\bottomrule
\end{tabular}}

\label{tab:rpr_personalization_interpretation}
\end{table}

\section{User-Level Personalization}
\label{app:user_personalization}
We use PersonalLLM for user-level adaptation.
Following prior work~\citep{bose2025lore, choi-etal-2025-copl}, we model each user as a weight vector over ten reward models. 
Adaptation and evaluation examples are then constructed from the highest- and lowest-ranked responses according to user-specific weights.
We sample 5 users from a Dirichlet distribution with $\alpha=0.3$, and use $n=50$ examples for adaptation. 
Evaluation is conducted on 200 synthetic examples.
Results are averaged over three independently sampled adaptation sets.

\begin{table*}[ht!]

    \centering
    \caption{\textbf{Personalization results on PersonalLLM.}
    Scores in parentheses denote the change after adaptation. 
    Higher scores are better.
    Sparse MoE also demonstrates strong performance in user-level adaptation.}
    \resizebox{\textwidth}{!}{
    \begin{tabular}{lcccccc}
    \toprule
    Method & User1 & User2 & User3 & User4 & User5 & Overall \\
    \midrule
     Single Reward & 93.83 & 94.83 & 91.50 & 94.67 & 87.17 & 92.40 \\
     HyRe & 95.50 (+0.83) & 95.17 (-0.33) & \textbf{93.00} (-0.17) & 95.33 (-0.50) & \textbf{89.50} (+1.00) & 93.70 (+0.17) \\
     Vanilla MoE & 94.83 (+0.00) & 94.67 (-0.83) & 90.67 (-2.00) & 94.67 (-1.33) & 88.83 (+0.50) & 92.73 (-0.74) \\
     MiCRo & 94.50 (-0.33) & 94.67 (-0.83) & 90.67 (-2.00) & 94.83 (-1.17) & 88.17 (-0.16) & 92.57 (-0.90)\\
     ArmoRM & 68.67  (+0.84)  & 72.67 (+1.84) & 71.50 (\textbf{+1.67}) & 72.17 (0.00) & 67.17 (\textbf{+1.85}) & 70.44 (+1.24)\\
    \midrule
    Sparse MoE & \textbf{95.83 (+4.50)} & \textbf{96.17 (+3.50)} & 92.67 (+1.34) & \textbf{97.33 (+3.33)} & 88.50 (-0.33) & \textbf{94.10 (+2.47)} \\
    \bottomrule
    \end{tabular}
    }
    
    \label{tab:user_personalization}
\end{table*}

Table~\ref{tab:user_personalization} shows that our sparse MoE reward model also achieves the strongest performance in user-level adaptation.
Consistent with the attribute-level adaptation results in Table~\ref{tab:rpr_personalization}, sparse MoE attains both the best post-adaptation performance and the largest improvements over the non-adapted model.
By contrast, the two MoE baselines, vanilla MoE and MiCRo, fail to adapt effectively to new users, with performance degrading after adaptation.
The static-weight ensemble model HyRe performs better than the MoE baselines, likely because its simpler adaptation mechanism is more robust in low-data settings.

\section{Personalized Alignment with Adapted Preferences}
\label{app:post_train}

\subsection{Sparse MoE for Downstream Post-Training}
To evaluate whether the adapted preferences can support personalized alignment of LLMs, we post-train policies using adapted sparse MoE reward models corresponding to different attributes from RPR.
Specifically, we select two attributes, creativity and scientific rigor, which exhibit improved performance after personalization.

For simplicity, we perform GRPO~\citep{grpo} post-training on prompts from the RPR training set for 1000 optimization steps.
We use an effective batch size of 8 and sample 4 rollouts per prompt.
Policies are initialized from \href{https://huggingface.co/meta-llama/Llama-3.2-3B-Instruct}{Llama3.2-3B-Instruct}.

During evaluation, we generate responses for 200 examples from the RPR test set and compare them against a baseline policy trained using the non-adapted reward model.
An LLM judge (Qwen3.6-27B) evaluates which response better reflects the target attribute while maintaining higher overall response quality.

\begin{table*}[h!]
\centering
    \caption{\textbf{Personalized alignment results compared against alignment using a universal reward model.}
    After personalization, the sparse MoE reward model can effectively guide policies toward the target attributes without negatively affecting overall generation quality.}
    \resizebox{0.8\linewidth}{!}{
    
    \begin{tabular}{l|cc|cc}
      \toprule
      \multirow{2}{*}{\makecell{Alignment\\Type}} & \multicolumn{2}{c|}{Creativity} & \multicolumn{2}{c}{Scientific Rigor} \\
      & Win Rate (Target Attribute, $\uparrow$) & Win Rate (Quality, $\uparrow$) & Win Rate (Target Attribute, $\uparrow$) & Win Rate (Quality, $\uparrow$)  \\
      \midrule
       Personalized   & \textbf{53.0} & \textbf{52.5} & \textbf{52.0} & \textbf{52.5} \\
       Universal & 45.5 & 45.5 & 44.5 & 45.0 \\
       \bottomrule
    \end{tabular}
    }

    \label{tab:post_train}
\end{table*}

Results in Table~\ref{tab:post_train} show that post-training policies with personalized reward models better align the policies with the target attributes compared to using a universal preference model.
This demonstrates that the personalization improvements of our sparse MoE reward models also generalize to the post-training phase.

\subsection{Personalized Reward Models for Pluralistic Alignment}

In this work, we mainly focus on one aspect of pluralistic alignment: how to \textit{represent} diverse human values in an interpretable and adaptable reward model. 
Optimizing policies toward these individualized value representations is a complementary research direction, and here we discuss how an individualized sparse MoE model can support pluralistic alignment in several ways.

First, one can train or fine-tune a separate policy for each user. In Appendix~\ref{app:post_train}, we show that an adapted sparse MoE model, trained with only a small number of examples (around 50), can be used as the optimization signal for GRPO to steer a policy toward the target preference direction. This approach is conceptually direct, but computationally expensive.

A more efficient option is to update and maintain only lightweight user-specific components, such as LoRA modules or user embeddings. Related test-time optimization methods~\citep{li2025testtime} may also reduce the need for full offline policy training.

The adapted reward model can also be used without explicit policy training, for example by guiding a shared universal policy at inference time. Reward-guided decoding methods such as PAD~\citep{chen2025pad}, DeAL~\citep{huang-etal-2025-deal}, and RAD~\citep{deng-raffel-2023-reward} show that reward models or customized reward functions can be used for decoding-time alignment without maintaining separate policies. In our setting, the individualized sparse MoE reward model could provide the user-specific reward signal.

Finally, generate-then-select methods, such as Best-of-N sampling~\citep{ichihara2025evaluation, sun2024fast}, offer another simple training-free alternative: a universal policy generates multiple candidate responses, and the individualized reward model selects the response that best aligns with the target user's preferences.

\section{Limitations}
\label{app:limitations}
Our study has certain limitations that should be acknowledged.

First, sparse MoE training introduces additional hyperparameters and tuning complexity. 
We ablate the number of experts and the regularization terms (Appendices~\ref{app:num_expert} and~\ref{app:ablation}) and show that one configuration remains effective across backbones and datasets (Appendix~\ref{app:generalization}). Our results therefore establish that interpretable preference components can be learned under suitable conditions; broader robustness across training configurations remains open.

Second, because the router conditions only on the prompt and not on the responses, router-based interpretations reveal the prompt-identifiable domains in which experts specialize rather than how they evaluate individual responses. Such task- and domain-level specialization is still meaningful for personalization, where users value performance differently across contexts, as our adaptation experiments show.
Recent work has explored imposing interpretability constraints on different components of MoE reward models, yielding distinct forms of interpretation~\citep{wang2026routingweightsfaithfulresponselevel}.

Third, sparse MoE slightly underperforms other MoE variants on universal preference accuracy. We stress that the two accuracies at stake are different: what sparse MoE gives up is 1.04 points relative to vanilla MoE on the population-average preference function (700K), while what it gains is an 18.63-point advantage over the same architecture on unseen preferences after adaptation (RPR). The trade-off is real but asymmetric: sparse MoE is most appropriate when interpretability and personalization are the objectives. Appendix~\ref{app:complete_results} analyzes the residual gap, which partly arises from misalignment between router assignments and expert strengths.

Finally, our study covers existing MoE reward models~\citep{wang-etal-2024-interpretable, shen-etal-2025-micro} trained with standard interpretability regularizers; how far expert structure could be pushed by purpose-built architectures or objectives remains open. Our contribution is diagnostic---what these models learn, why conventional training leaves it entangled, and what that entanglement costs---together with the evaluation protocol that supports such claims. We hope both facilitate further work on interpretable preference modeling.
\end{document}